\RequirePackage{fix-cm}
\documentclass[smallextended]{svjour3}       
\smartqed  
\usepackage{graphicx}
\usepackage{amsmath}
\usepackage{amssymb}
\usepackage{multirow}
\usepackage{subfigure}
\usepackage{float}
\begin{document}

\title{Detailed 3D Human Body Reconstruction from Multi-view Images Combining Voxel Super-Resolution and Learned Implicit Representation
}


\author{Zhongguo Li        \and
        Magnus Oskarsson \and  Anders Heyden
}


\institute{Zhongguo Li \at
              S{$\ddot{\rm{o}}$}lvegatan 18, Lund \\
              Tel.: +46-738944299\\
              \email{zhongguo.li@math.lth.se}           
           \and
           Magnus Oskarsson
           \and
           Anders Heyden
           \at
           S{$\ddot{\rm{o}}$}lvegatan 18, Lund
}

\date{Received: date / Accepted: date}

\maketitle

\begin{abstract}
The task of reconstructing detailed 3D human body models from images is interesting but challenging in computer vision due to the high freedom of human bodies. In order to tackle the problem, we propose a coarse-to-fine method to reconstruct a detailed 3D human body from multi-view images combining voxel super-resolution based on learning the implicit representation. Firstly, the coarse 3D models are estimated by learning an implicit representation based on multi-scale features which are extracted by multi-stage hourglass networks from the multi-view images. Then, taking the low resolution voxel grids which are generated by the coarse 3D models as input, the voxel super-resolution based on an implicit representation is learned through a multi-stage 3D convolutional neural network. Finally, the refined detailed 3D human body models can be produced by the voxel super-resolution which can preserve the details and reduce the false reconstruction of the coarse 3D models. Benefiting from the implicit representation, the training process in our method is memory efficient and the detailed 3D human body produced by our method from multi-view images is the continuous decision boundary with high-resolution geometry. In addition, the coarse-to-fine method based on voxel super-resolution can remove false reconstructions and preserve the appearance details in the final reconstruction, simultaneously. In the experiments, our method quantitatively and qualitatively achieves the competitive 3D human body reconstructions from images with various poses and shapes on both the real and synthetic datasets. 
\end{abstract}

\section{Introduction}
\label{intro}
Recovering detailed 3D human body models from images attracts much attention because of its wide applications in movie industry, animations, and Virtual/ Augmented Reality. However, inferring 3D objects from 2D images is a challenging task in computer vision due to the ambiguity of reprojection from 2D to 3D space. The high freedom of the human body in real scenes further increases the difficulty of the task. Although multi-view systems~\cite{joo2017panoptic} and laser scanning systems~\cite{xu2019unstructuredfusion} are now able to reconstruct accurate 3D human bodies, these systems remain inconvenient for common users because they are often hard to deploy and expensive. Thus, estimating 3D human bodies from images is more attractive and many approaches have provided possible directions to tackle the problem from advocating the pre-defined parametric human body as template to recent deep learning based route. 

Traditionally, 3D human body reconstruction from RGB images mainly depends on the pre-defined parametric human body models. From simple geometric primitives~\cite{sigal2004tracking} to data-driven models~\cite{Anguelov_2005,Loper_2015smpl}, parametric human body models play an important role in human related research. The main idea of the route is to fit the parametric human body model to some prior information including the skeleton, 2D joint points and the silhouettes~\cite{Balan_2007detailed,Bogo_2016keep,Alldieck_2018video}. Such methods have been used for human motion tracking and 3D pose estimation successfully. However, due to the missing detailed appearance on the most parametric human bodies such as clothes and facial expression, the results of these methods are often unclothed, which cannot satisfy the requirements of the realism in many applications.

Benefiting from the great success of deep learning in many computer vision tasks, 3D human body reconstruction from images based on deep learning also has achieved some progress recently. During the past several years, convolutional neural networks (CNN) have shown impressive performance on 2D/3D human pose estimation~\cite{pishchulin2016deepcut,newell2016stacked,alp2018densepose} and human body segmentation~\cite{xia2017joint,he2017mask}. Therefore, some methods automatically estimated 3D human body model from images by fitting the parametric human body to prior cues like the 2D or 3D joint points of human body and silhouettes which can be estimated by the CNN~\cite{Bogo_2016keep,Huang_2017towards,Alldieck_2018video,Xu_2018monoperfcap}. Since the poses and silhouettes comprise sparse information, directly inferring the pose and shape of a parametric human body model from the full image through the CNN become another useful route and has achieved impressive performance~\cite{Kanazawa_2018end,Pavlakos_2018learning,Pavlakos_2019texturepose,Kolotouros_2019convolutional,Kolotouros_2019learning}. However, the 3D human body models obtained by these methods are still unclothed. Recently, many approaches came up with a refining process based on CNN on the parametric human body to add clothes on the naked 3D human body model. The refining process includes the image texture translation~\cite{alldieck2019tex2shape}, inferring the surface normals~\cite{saito2020pifuhd} and volumetric regression~\cite{zheng2019deephuman}. Through refining the parametric human body model, these methods can obtain some details including the clothes and hair on the final 3D model. However, these methods require that the parametric human body model has the accurate pose with the observed human body because the final estimation will be affected seriously if the prior information is not predicted correctly.

Recently, deep learning on 3D reconstruction like point clouds or voxels from images for some general objects has gained popularity. Explicit volumetric representations are straightforward for learning to infer 3D objects from RGB images~\cite{choy20163d,Kar_2017learning,wang2018pixel2mesh,fan2017point}. Due to the limitation of memory, these methods can only produce low-resolution 3D objects (e.g. $32^3$ or $64^3$ number of voxels). Even though some methods reduce the memory footprint through adaptive representations such as octrees, the final resolutions are sill relatively small (e.g. $256^3$)~\cite{riegler2017octnet}. In addition to this, these results are always discrete, which results in the missing of many details on the surface. In contrast to explicit representations, implicit function for 3D model representation in deep learning shows impressive performance~\cite{park2019deepsdf,Mescheder_2019occupancy,Chen_2019learning,chibane20ifnet} and are attracting much attention. Compared to learning the explicit volumetric representation, learning an implicit function to represent 3D shape can be implemented in a memory efficient way, especially for the training process. Another advantage of implicit representation is that the 3D model can be decided by the continuous decision boundary, which allows a high-resolution 3D model. Considering the advantages, there are some methods based on learning implicit function to reconstruct detailed 3D human body from images~\cite{Huang_2018deep,saito2019pifu,saito2020pifuhd}. However, these methods may still produce some false reconstruction on the final 3D model.  

\begin{figure}
  \includegraphics[width=\linewidth]{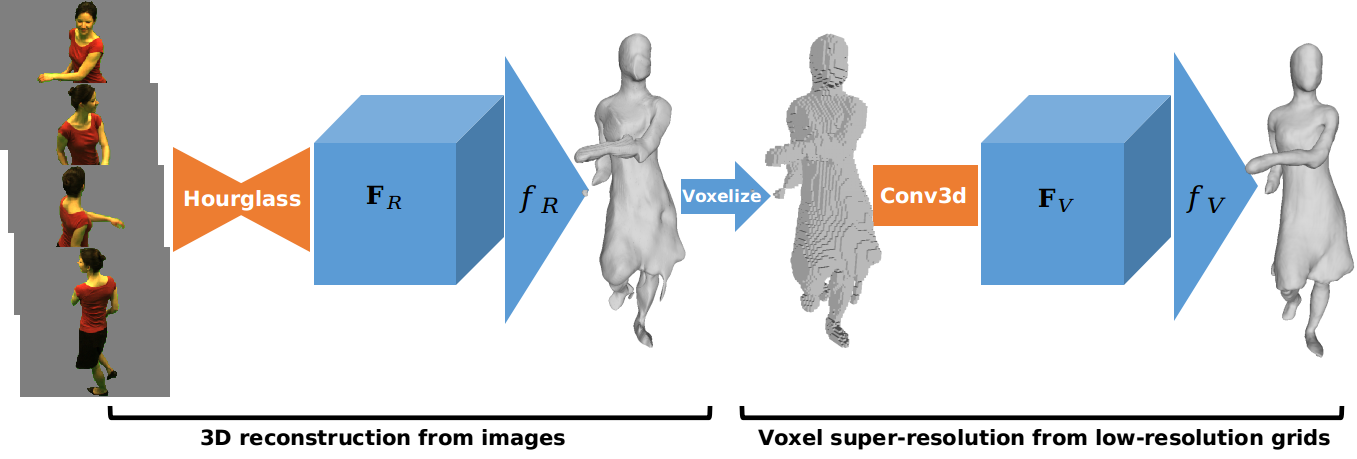}
   \caption{The pipeline of our method. It consists of 3D reconstruction from images and voxel super-resolution from low-resolution grids. The 3D reconstruction from images estimates a coarse 3D human body model. After voxelizing coarse model to a low-resolution grid, the voxel super-resolution refines the low-resolution grid to obtain detailed model.}
\label{fig:overview}
\end{figure}

In this paper we propose a novel method to estimate a detailed 3D human body model from multi-view images, through learning an implicit representation. Our method works in a coarse-to-fine manner, and thus, consists of two parts: (1) inferring the 3D human body model from multi-view images, and (2) voxel super-resolution from low-resolution voxel grids obtained by (1). In both of the two parts, we attempt to learn an implicit function to represent the 3D models. For the reconstruction of a 3D human body from multi-view images in (1), the structure of multi-stage hourglass networks is designed to produce multi-scale features and a fully connected neural network predicts the occupancy values of the features to implicitly represent 3D models. Through training the above model, the coarse 3D models can be estimated from multi-view images. Then, low-resolution grids can be generated by voxelizing the coarse models. Taking the low-resolution grids as input, a multi-stage 3D CNN is built to produce multi-scale features and a fully connected neural network is also utilized to predict the occupancy values of the features. The final 3D model is generated by the implicit representation through refining the coarse model by voxel super-resolution. Our method is summarized in Fig.~\ref{fig:overview}.

Our method differs from previous work in three aspects. Firstly, it is a coarse-to-fine method  combining 3D reconstruction from multi-view images and voxel super-resolution into one route to infer 3D human body models. The 3D reconstruction from images produces a coarse result and the voxel super-resolution refines the coarse result to generate a final detailed 3D reconstruction. Secondly, the implicit representation for the 3D model is used both in image based 3D reconstruction and voxel super-resolution, which is memory efficient for training and can produce high resolution geometry through extracting a continuous decision boundary. Finally, the multi-scale features are extracted from multi-view images and low-resolution voxel-grids for coarse reconstruction and refining the models, respectively. The multi-scale features are able to fully encode the local and global spatial information of the pixels in the images and the voxels in the low resolution voxel grids. 

The paper is organized as follows. The introduction and related work of our method are presented in Section 1 and Section 2, respectively. The following Section 3 describes the detailed coarse-to-fine structure of our method and the implementation details including the 3D model reconstruction from multi-view images and voxel super-resolution. In Section 4, some quantitative and qualitative experiments are illustrated to evaluate the performance of our method. Finally, the conclusion and future work are stated in Section 5. 
\section{Related work}
\label{sec:rw}
We summary the related work on 3D human body reconstruction from images and 3D vision based on deep learning in this section. There are three parts in the section: (1) Optimization based methods; (2) Parametric human body model based regression, and (3) Non-parametric human body model based regression.

\textbf{Optimization based methods.} The classic route to recover 3D human body models from an image is to fit a template such as SCAPE~\cite{Anguelov_2005} or SMPL~\cite{Loper_2015smpl} to prior cues. SCAPE, which was a data-driven parametric human body model to represent human pose and shape, was learned from 3D human body scans~\cite{Anguelov_2005}. Some methods fitted SCAPE to the silhouettes and joint points from observed images to recover human pose and shape~\cite{Balan_2007detailed,sigal2008combined,Guan_2009estimating}. With the emergence of Kinect, the depth images were also used for fitting the SCAPE~\cite{Weiss_2011home,Bogo_2015detailed,Liu_2016template}. With the success of deep learning on human pose estimation~\cite{newell2016stacked,martinez2017simple,alp2018densepose,Cao_2019openpose}, the joint points can be obtained automatically with high accuracy. In~\cite{Bogo_2016keep}, an automatic method for 3D human body estimation was proposed through fitting a novel parametric human body model called SMPL~\cite{Loper_2015smpl} to the 2D joint points predicted by deep learning~\cite{pishchulin2016deepcut}. Then, more methods turned to use SMPL or pre-scanning models for human body reconstruction based on 3D joint points, multi-view images, video and silhouettes\cite{Huang_2017towards,Alldieck_2018video,Xu_2018monoperfcap,Li_2019parametric,habermann2019livecap}. These methods tried to build better energy function based on various prior cues and the 3D human body was estimated by optimizing the energy function. Although the optimization based methods were classic, the estimated 3D human body was always unclothed due to the limitation of parametric human body, which limited its realism. 

\textbf{Parametric human body model based regression.} Since deep learning has achieved impressive performance on many computer vision tasks, it also attracts much attention on 3D human body estimation through regressing the parametric human body model. In the beginning, the shape parameters of SCAPE were regressed from silhouettes to estimate 3D human body model in~\cite{Dibra_2016hs,Dibra_2017human}, which can only handle the standing pose or very simple poses. In ~\cite{tan2017indirect}, the shape and pose of the SMPL model were regressed through the images and the corresponding SMPL silhouettes. Instead of using silhouettes, the authors proposed to take the whole image as the input of the CNN to regress the pose and shape parameters of the SMPL model thorough building the loss function about the joint points~\cite{Kanazawa_2018end}. Since then, many improved methods were proposed through designing novel network structure or using more constraints on the loss function~\cite{Pavlakos_2018learning,Kolotouros_2019convolutional,Pavlakos_2019texturepose,Kolotouros_2019learning,Kanazawa_2019learning,Liang_2019shape,Kocabas_2020vibe}. Pavlakos et al.~\cite{Pavlakos_2018learning} combined joint points and silhouettes in the loss function to better estimate the shape. There were some other approaches in which various cues were used for building sufficient loss function to train the network including the mesh~\cite{Kolotouros_2019convolutional}, 
the texture~\cite{Pavlakos_2019texturepose}, the multi-view images~\cite{Liang_2019shape}, the optimized SMPL model~\cite{Kolotouros_2019learning} and the video~\cite{Kanazawa_2019learning,Kocabas_2020vibe}. Although these methods can infer the pose and shape of SMPL model very well, they still obtained unclothed human body models. 
In order to model the detailed appearance, some method attempt to refine the regressed SMPL model to obtain the detailed 3D model~\cite{alldieck2018detailed,varol2018bodynet,zhu2019detailed,lazova2019360,alldieck2019tex2shape,zheng2019deephuman,onizuka2020tetratsdf,huang2020arch}. In~\cite{alldieck2018detailed}, after estimating the pose and shape of SMPL model, the authors used shape from shading and texture translation to add the details to SMPL like face, hairstyle, and clothes with garment wrinkles. They also proposed some improved methods to obtain better results~\cite{lazova2019360,alldieck2019tex2shape}. In addition to the texture, the explicit representation of 3D human body model were also used in detailed reconstruction. BodyNet~\cite{varol2018bodynet} added the volume loss function to better estimate the pose and shape of SMPL. DeepHuman~\cite{zheng2019deephuman} refined the appearance of volumetric SMPL model through transferring the image normal to the volumetric SMPL. In~\cite{onizuka2020tetratsdf}, a novel tetrahedral representation for SMPL model was used and the detailed model was obtained by learning the sign distance function of tetrahedral representation. Another recent work also refined the normal and color of image to the estimated SMPL model~\cite{huang2020arch} from single image. 

\textbf{Non-parametric human body model based regression.} Recently, deep learning also achieved some success on reconstruction of 3D objects from images without relying on any parametric models. Some methods tried to extract coarse 3D information from 2D images and attempted to refine the 3D information through deep neural network such as volume, visual hull, depth images~\cite{jackson20183d,Huang_2018deep,gilbert2018volumetric,gabeur2019moulding,natsume2019siclope}. Jackson et al.~\cite{jackson20183d} reconstructed 3D geometry of humans through training an end-to-end CNN to regress the volumes which were provided in the training dataset. In~\cite{gilbert2018volumetric}, a coarse model was obtained though Visual Hull from sparse view images and the coarse model was refined by a deep neural network. Natsume et al.~\cite{natsume2019siclope} generated multi-view silhouettes through deep learning from single image and proposed a deep visual hull to infer the detailed 3D models based on the estimated silhouettes. Huang et al.~\cite{Huang_2018deep} estimated detailed models by deciding if a spatial point inside or outside of 3D mesh through classifying the features extracted by the CNN. Gabeur et al.~\cite{gabeur2019moulding} estimated the visible and invisible point clouds of the human body from image through deep learning and the full detailed body can be formed by the point clouds. Instead of inferring 3D information from images, some other methods gained popularity to reconstruct general 3D models directly from images with explicit representation such as voxels and point cloud~\cite{choy20163d,Kar_2017learning,wang2018pixel2mesh,fan2017point}. Due to the limitation of resolution of an explicit representation, implicit representation of 3D models based on deep learning have been used for reconstruction of general objects~\cite{Kar_2017learning,Mescheder_2019occupancy,Chen_2019learning,chibane20ifnet}. Inspired by the idea, some methods only for detailed 3D human body reconstruction also proposed based on learning implicit representation. Saito et al.~\cite{saito2019pifu} extracted the pixel-aligned features from images through end-to-end networks. Associating the depth of pixel, the implicit representation can be learned from the features. The method can produce the high-resolution detailed 3D human body including the facial expression, clothes and hair can be estimated from by the above methods. However, there existed many errors on the estimation because only 2D images were used. An improved method called PIFuHD~\cite{saito2020pifuhd} was proposed to reconstruct high-resolution detailed 3D human body from images through introducing image normal to PIFu. The coarse-to-fine methods could obtained more accurate reconstruction because more cues were used for the reconstruction.
\section{Method}
In this section the details of our method are described. We firstly introduce the background of implicit function to represent the 3D shape. Then, we present the 3D human body reconstruction from multi-view images through learning the implicit representation. Afterwards, an implicit representation based network for voxel super-resolution is presented to refine the 3D human body model obtained from the multi-view images. Finally, the implementation details of our method are introduced. 
\subsection{Learning an implicit function for 3D models}
For 3D reconstruction based on deep learning, implicit functions to represent 3D shape is memory efficient for training. Instead of storing all voxels of the volume in an explicit volumetric representation, an implicit function for 3D representation assigns the signed distance or occupancy probability to a spatial point to decide if the point lies inside or outside of the 3D mesh. From this the 3D mesh can be extracted by a level set surface. In our method, we use occupancy probability as the output of the implicit function. Given a spatial point and a water-tight mesh, the occupancy function is defined as:
\begin{equation}
    f(X):=x, X\in\mathbb{R}^{3}, x\in \{0,1\},
\end{equation}
where $X$ is the 3D point and $x$ is the value of occupancy function for $X$. The value of $x$ indicates if $X$ lies inside (0) or outside (1) of the mesh. The 3D mesh can be implicitly represented and generated by the level set of $f(X)=0.5$.

For 3D reconstruction based on learning implicit representation, the key problem is to learn the occupancy function $f(\cdot)$. More specifically, a deep neural network encodes 3D shape as a vector $\mathbf{v} \in \mathcal{V} \subset \mathbb{R}^m$, and then, the occupancy function takes the vector as input to decide the value of the 3D point, i.e.,
\begin{equation}
    f(\mathbf{v},X):\mathcal{V}\times \mathbb{R}^3 \mapsto [0,1].
\end{equation}
As long as $f(\cdot)$ can be learned, the continuous occupancy probability field of a
3D model can be predicted and the 3D model can be extracted by the iso-surface of the field through the classic Marching Cubes algorithm.

In PIFu~\cite{saito2019pifu}, the authors presented a pixel-aligned implicit function for high-resolution 3D human body reconstruction. It is defined as:
\begin{equation}
f(F(\pi(X)),z(X)):\mathcal{V}\times \mathbb{R} \mapsto [0,1],  
\end{equation}
where $F(\cdot)$ is the feature grids of CNN, $\pi(X)$ is the projection of $X$ on the image plane by $\pi$ and $z(X)$ is the depth of $X$. 
PIFu showed impressive performance on detailed reconstruction of human bodies for fashion poses, for instance, walking and standing. However, the features extracted by multi-stage networks from input images have the same scale, which may result in the missing of some details. In addition, for some complicated poses, only using 2D images may result in false reconstructions. Aiming at the above two drawbacks, we propose two improvements. On one hand, the multi-scale features are extracted in both 3D reconstruction from images and voxel super-resolution. On the other hand, the voxel super-resolution refines the coarse 3D models to reduce false reconstructions. 

The outline of our method is shown as Fig.~\ref{fig:overview}. It has two parts: (1) 3D reconstruction from images; and (2) Voxel super-resolution from low-resolution grids. The details of the two parts are presented in the following sections.  
\subsection{MF-PIFu}
\begin{figure}[htbp]
  \includegraphics[width=0.7\linewidth]{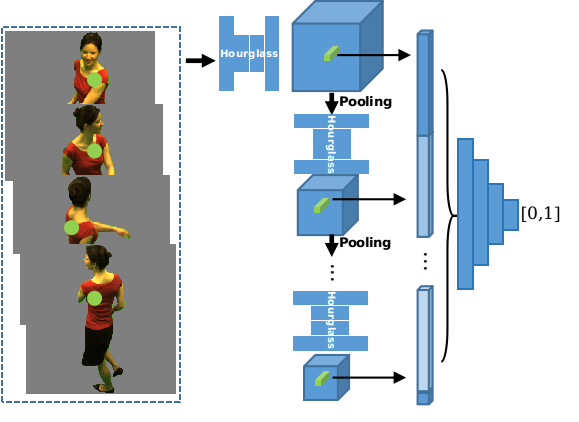}
   \caption{The structure of MF-PIFu to learn the implicit representation of 3D human body model. Multi-stage hourglass networks are used for multi-scale feature extraction and a fully connected neural network predicts the occupancy value of the feature.}
\label{fig:part1}
\end{figure}

The method for 3D reconstruction from multi-view images is inspired by PIFu~\cite{saito2019pifu}. The difference is that we extract Multi-scale Features from multi-view images through multi-stage hourglass networks. Therefore, we call our method as MF-PIFu and the architecture of MF-PIFu is shown in Fig.~\ref{fig:part1}.  

Given images with $N$ views $I_i,i=1,...,N$, multi-stage hourglass networks encode the images as feature grids $\mathbf{F}_R^{(j)}, j=1,...,M$ where $M$ is the number of hourglass networks. We denote the multi-stage hourglass network as $g_R(\cdot)$. Then, for the $i$-th image $I_i$, its multi-scale feature grids are defined as:
\begin{equation}
    g_R(I_i):=\mathbf{F}_R^{(i,1)}, ..., \mathbf{F}_R^{(i,M)},
\end{equation}
where the feature grids $\mathbf{F}_R^{(i,1)}, ..., \mathbf{F}_R^{(i,M)}$ have different scales and the $j$-th grid $\mathbf{F}_R^{(i,j)}$ belongs to feature space $\mathcal{F}_j^{C\times K\times K}$. $C$ is the depth of feature grid and $K$ is the width and height of the feature grid. In our method, $C$ is kept constant (e.g. 256) and $K$ deceases as $2^{j-1}$ for the $j$-th hourglass network. Before the $\mathbf{F}_R^{(i,j-1)}$ is fed into the $j$-th hourglass nwtwork, we use a max-pooling layer to downsample $\mathbf{F}_R^{(i,j-1)}$. Through this max-pooling layer, the multi-scale feature grids can be generated by the multi-stage hourglass networks. For the pixel $x$ in the image $I_i$, the feature vector in $\mathbf{F}_R^{(i,j)}$ can be obtained at the corresponding location through interpolation, which is denoted as $\mathbf{F}_R^{(j,1)}(x)\in\mathcal{F}_j^{C}$. 

After getting the multi-scale features, we need to query the multi-scale features, i.e., predict the occupancy value. The prediction is defined by a fully connected neural network which is defined as $f_R(\cdot)$. Similar to PIFu, not only the features are used for prediction, but also the depth of the corresponding pixel is also used. The multi-scale features and the depth form new feature vector for prediction. For the pixel $x$ in the image $I_i$, we define the new feature vector as $\mathbf{F}_R^{(i)}(x)=\{\mathbf{F}_R^{(i,1)}(x), ..., \mathbf{F}_R^{(i,M)}(x),z(x)\}\in\mathcal{F}_1^{C}\times ...\times\mathcal{F}_M^{C}\times \mathbb{R}$. The fully connected neural network takes into the feature vector to predict the occupancy value of $x$:
\begin{equation}
    f_R(\mathbf{F}_R^{(i)}(x)):\mathcal{F}_1^{C}\times ...\times\mathcal{F}_M^{C}\times \mathbb{R} \mapsto [0,1].
\end{equation}
In contrast to PIFu, we form the features from each stage and the depth as a new feature vector. This new feature encodes both the local and global information of the pixels. The feature grids at an early stage encode more local information, while the feature grids at the last stage represent the global information. Associating the depth information, the new features encode more information than the features used in PIFu, and thus, it is more reliable for prediction of occupancy value. 
\begin{figure}[htbp]
  \includegraphics[width=0.7\linewidth]{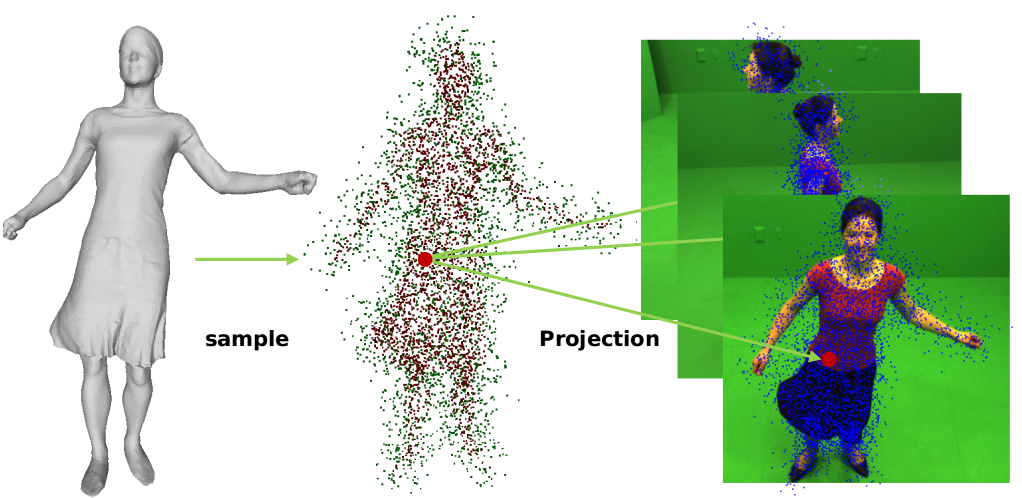}
   \caption{Sampling 3D points from 3D model and projecting the points to multi-view images.}
\label{fig:proj}
\end{figure}

To train $g_R(\cdot)$ and $f_R(\cdot)$ from multi-view images $I_i,i=1,...,N$, the pairs $\{I_i,\mathcal{S}\}$ are required in which $\mathcal{S}$ is the corresponding ground truth of 3D model for the multi-view images $I_i$. As shown in Fig.~\ref{fig:proj}, 3D spatial points $X_i,i=1,...,K$ are sampled from the 3D model $\mathcal{S}$ and are added random displacements with normal distribution $\mathcal{N}(0,\sigma)$ on the points. This means that the points to be queried are $\hat{X}_i=X_i+n_i$ where $n_i\sim\mathcal{N}(0,\sigma)$. The binary occupancy values of the points $o(\hat{X}_i)$ can be obtained according to the location of $\hat{X}_i$. If $\hat{X}_i$ lies in $\mathcal{S}$, $o(\hat{X}_i)=0$. Otherwise, $o(\hat{X}_i)$ is 1. The points $\hat{X}_i$ are projected onto the multi-view images through the given camera parameters. The corresponding pixel of point $\hat{X}_j$ on the $i$-th image is $x_{ij}=\pi_i(\hat{X}_j)$. Then, the loss function for the pair $\{I_i,\mathcal{S}\}$ can be defined as:
\begin{equation}
L_R =\sum_{i=1}^{N}\sum_{j=1}^{K}\|f_R(\mathbf{F}_R^{(i)}(x_{ij}))-o(X_j)\|.
\end{equation}
In the above loss function, $\mathbf{F}_R^{(i)}(x_{ij})$ is the multi-scale features of pixel $x_{ij}$ which is the projection of 3D point $\hat{X}_j$ on the $i$-th view image. This loss function is defined based on the multi-view images jointly, which can predict the occupancy values more accurately. Through minimizing the loss function, $g_R(\cdot)$ and $f_R(\cdot)$ can be trained end-to-end.
\subsection{Voxel Super-Resolution}
\begin{figure}
  \includegraphics[width=0.7\linewidth]{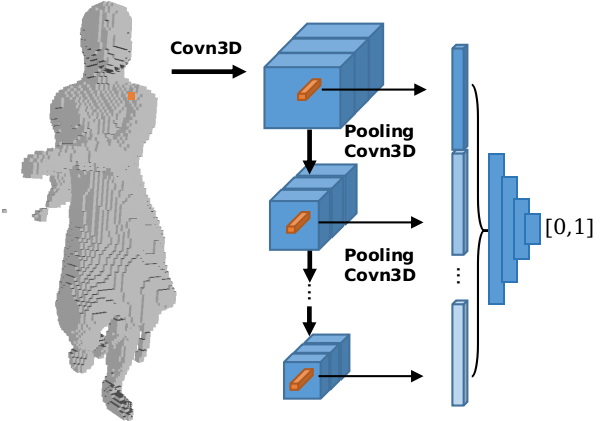}
   \caption{The structure of voxel super-resolution based on learning implicit representation. Multi-stage 3D convolutional layers are used for extracting the multi-scale features from low-resolution grid. A fully connected neural network is used for predicting occupancy value of features. }
\label{fig:part2}
\end{figure}

The 3D models recovered by MF-PIFu are still coarse because MF-PIFu only relies on 2D images. We observe two problems in the estimated 3D models by MF-PIFu. The first one is that the surface of the 3D model is not smooth due to the multi-view effect. The second one is that some extra unnecessary parts are reconstructed on the models due to the false classification of some voxels. In order to overcome the problems, we propose the voxel super-resolution (VSR) to refine the coarse 3D models of MF-PIFu. As shown in Fig.~\ref{fig:part2}, our VSR method also uses a multi-scale structure for feature extraction and implicit representation for the 3D model. In contrast to MF-PIFu which uses images as input, the input of VSR is a low resolution voxel grid which is produced by the voxelization of the 3D model of MF-PIFu. 
Suppose the 3D model estimated by MF-PIFu is $\hat{\mathcal{S}}$ which is stored as the voxel positions. The voxelization of $\hat{\mathcal{S}}$ can produce a low resolution grid as $\mathcal{V}\in \mathbb{R}^{N\times N \times N}$ (e.g. N=128). Then, as shown in Fig.~\ref{fig:part2}, 3D convolution kernels are utilized to extract 3D feature grids from $\mathcal{V}$. We recursively use $n$ 3D convolution layers to generate the multi-scale feature grids $\mathbf{F}_V^{(1)}, ...,\mathbf{F}_V^{(n)}$. The resolution of the $k$-th feature grid is $N/(2^{k-1})$, i.e., $\mathbf{F}_V^{(k)}\in \mathcal{F}_k^{K\times{K}\times{K}}$ where $K=N/(2^{k-1})$. The resolution of the feature grids decreases with the depth of the network. We denote the 3D convolution neural network for VSR as $g_V(\cdot)$ and the multi-scale features can be generated as:
\begin{equation}
    g_V(\mathcal{V}):=\mathbf{F}_V^{(1)}, ...,\mathbf{F}_V^{(n)}.
\end{equation}
The feature grid at the early stage encodes more local information such as the shape details, while the feature grid at the late stage captures the global information of the voxel grid because of the large receptive fields at the late stage. 

For a voxel $\mathbf{v}\in \mathcal{V}$, its corresponding multi-scale feature is formed by the features from $\mathbf{F}_V^{(1)}, ...,\mathbf{F}_V^{(n)}$. Since the feature grid is discrete, the feature of voxel $\mathbf{v}$ in $\mathbf{F}_V^{(k)}$ is extracted by trilinear interpolation and is denoted as $\mathbf{F}_V^{(k)}(\mathbf{v})$. The multi-scale feature for the voxel $\mathbf{v}$ is 
\begin{equation}
    \mathbf{F}_V(\mathbf{v})=\{\mathbf{F}_V^{(1)}(\mathbf{v}),...,\mathbf{F}_V^{(n)}(\mathbf{v})\},
\end{equation}
where $\mathbf{F}_V(\mathbf{v})\in\mathcal{F}_1\times...\times{\mathcal{F}_n}$. After obtaining the multi-scale feature for a voxel $\mathbf{v}$, we also use a fully connected network to classify the multi-scale feature and and we denote it $f_{V}(\cdot)$. The fully connected network predicts the occupancy value of the multi-scale feature of $\mathbf{F}_V(\mathbf{v})$:,
\begin{equation}
    f_V(\mathbf{F}_v(\mathbf{v})):\mathcal{F}_1 ...\times{\mathcal{F}_n}\mapsto\in[0,1]
\end{equation}
This fully connected neural network classifies the voxel based on the multi-scale feature if the corresponding point lies inside or outside of 3D mesh. The implicit representation enables to produce a continuous surface. Besides, since multi-scale feature encodes both the local and global information, the 3D model after super-resolution can keep the global shape and preserve details of the shape.
\begin{figure}[htbp]
  \includegraphics[width=0.7\linewidth]{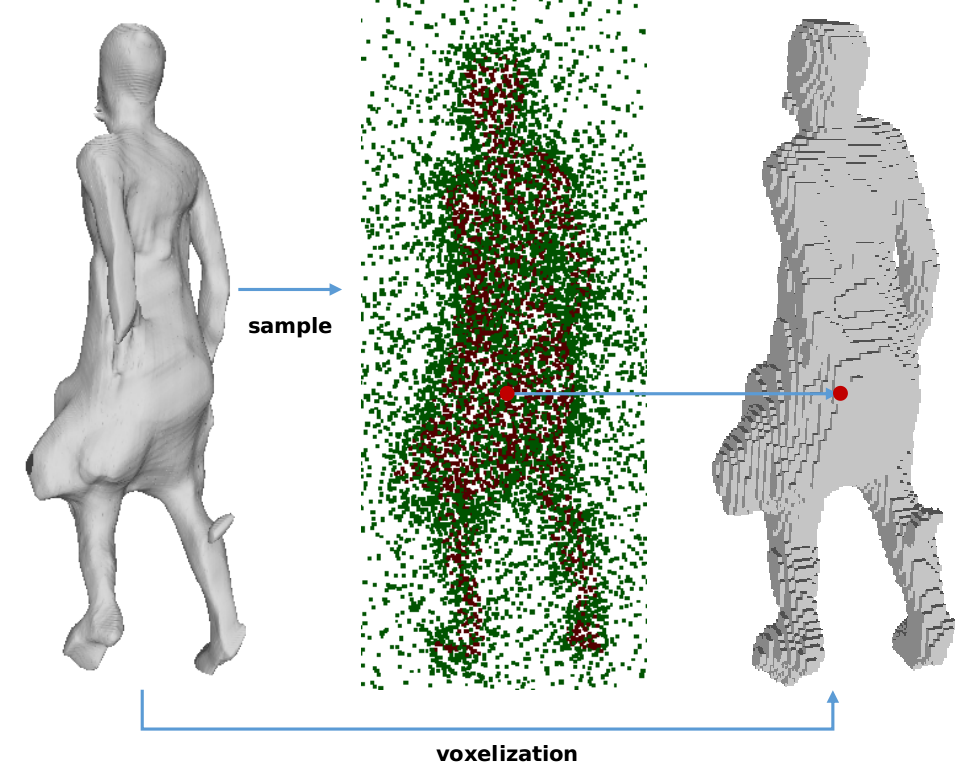}
   \caption{Sampling 3D points from 3D model estimated by MF-PIFu and the voxelization of the 3D model estimated by MF-PIFu (The resolution is $128^3$). The 3D points can be indexed by the grid coordinates in the low-resolution grid.}
\label{fig:voxelization}
\end{figure}

In order to train the $g_V(\cdot)$ and $f_V(\cdot)$ from low-resolution voxel grids $\mathcal{V}$, the 3D model $\mathcal{\hat{S}}$ estimated by MF-PIFu and its ground truth $\mathcal{S}$ are given as a pair $\{\mathcal{\hat{S}},\mathcal{S}\}$. The input low-resolution voxel grids are generated by voxelizing $\mathcal{\hat{S}}$. Instead of sampling points from $\mathcal{S}$, we sample N points $\mathbf{v}_i,i,...,N$ on the surface of $\mathcal{\hat{S}}$ and add random displacements with normal distribution $n_i\sim N(0,\sigma)$ to these points, i.e., $\hat{\mathbf{v}}_i=\mathbf{v}_i+n_i$. Here we take the same strategy as~\cite{chibane20ifnet} to generate points to be queried, i.e., 50\% points $\mathbf{v}_i$ are added random displacements with small $\sigma_{\min}$ and the other 50\% points $\mathbf{v}_i$ are added random displacements with large $\sigma_{\max}$. During the voxelization, the grid coordinates of the points $\hat{\mathbf{v}}_i$ in the low-resolution voxel grids $\mathcal{V}$ can be indexed and we denote it as $\rho(\hat{\mathbf{v}}_i)$. One example of sampling points and voxelization to a $128^3$ grid is shown in Fig.~\ref{fig:voxelization}. According to whether the point lies inside or outside of the ground truth 3D model $\mathcal{S}$, the binary occupancy value of the points $\hat{\mathbf{v}}_i$ can also be obtained as $o(\hat{\mathbf{v}}_i)$. We can do this because the estimated 3D model by MF-PIFu has been close to the ground truth. Through sampling the points on the estimated 3D model, the occupancy values of the points are reliable to do the voxel super-resolution. After getting the occupancy value of the points, the loss function for training the model of voxel super-resolution can be defined as:
\begin{equation}
\begin{aligned}
    L_{VSR}&=\sum_{i=1}^{N}\|f_{V}(g_V(\rho(\mathbf{\hat{v}}_i)))-o(\mathbf{\hat{v}}_i)\| \\
           &=\sum_{i=1}^{N}\|f_{V}(\mathbf{F}_V(\rho(\mathbf{\hat{v}}_i)))-o(\mathbf{\hat{v}}_i)\|.
\end{aligned}
\end{equation}
In the loss function, multi-scale features are used, and thus, the local and global information of the low-resolution voxel gird are encoded, which can preserve the details and the global shape simultaneously. We use standard cross-entropy loss function to measure the loss between the prediction and ground truth. Through minimizing the loss function $L_{VSR}$, the multi-stage 3D convolutional neural networks and the fully connected network are trained. 

\subsection{Implementation Details}
As shown in Fig.~\ref{fig:overview}, our model is a coarse-to-fine architecture in which MF-PIFu reconstructs coarse 3D models from multi-view image and VSR refines the coarse models to produce models with high accuracy. In this section the implementation details about the network structure, training and testing of our method are presented. 

\textbf{Network structure of MF-PIFu.} We use four stages of hourglass networks to generate multi-scale features and four layers in the fully connected neural network for prediction of occupancy value. For the extraction of multi-scale features, the input of the networks is the multi-view images (e.g. four views in the most of our experiments) which have removed backgrounds and are cropped to $256\times256$. The hourglass network consists of two convolutional layers and two deconvolutional layers to generate pixel-aligned feature maps. Max pooling is used for downsampling the feature maps. The output feature grids of each hourglass network has the size of $256\times 128\times\ 128$, $256\times 64\times\ 64$, $256\times 32\times\ 32$, and $256\times 16\times\ 16$. The fully connected network has four convolutional layers and the number of neurons in each layer is $(1024,512,128,1)$. The input feature of the fully connected layer has size 1025 because the multi-scale features also consider the depth of queried pixel.

\textbf{Training for MF-PIFu.} During the training, the batch size of input images is 4 and the model is trained for 12 epochs. In addition, 10,000 points are sampled from the ground truth of 3D mesh and they are added normally random noise with $\sigma=5\ cm$. These points are used for prediction of the occupancy value to build the loss function. The Mean Square Error (MSE) is used for building the loss function. The RMSProp algorithm with initial learning rate $0.001$ is used for updating the weights of the networks and the learning rate decreases by a factor of 0.1 after 10 epochs. It takes about 7 hours for training on our dataset.

\textbf{Network structure of VSR.} The architecture for VSR has the multi-stage 3D convolutional layers for generating multi-scale features from low resolution voxel grids and the fully connected neural network to predict the occupancy value of the multi-scale features. The input of the 3D convolution neural network is the low resolution voxel grids which have the size $128^3$. We use 5 stage 3D convolutional layers and the max pooling is used for downsampling the feature maps. The output feature grid of each convolution block has size of $16\times(128\times128\times128)$, $32\times(64\times64\times64)$, $64\times(32\times32\times32)$, $128\times(16\times16\times16)$, $128\times(8\times8\times8)$. Therefore, the input feature vector of the fully connected nerual network has $368$ elements. The fully connected neural network for predicting the occupancy value consists of four convolutional layers and the number of neurons in each layer is (256,256,256,1). 

\textbf{Training for VSR.} The low-resolution voxel grids for training the VSR is generated by the coarse 3D models estimated by MF-PIFu through voxelization. The input low-resolution voxel grids have resolution $128^3$. We sample 10,000 points from the coarse 3D models, in which 50\% of the points are added normal distribution displacements with $\sigma_{\max}=15\ cm$ and the other 50\% of the points are added normal distribution displacements with $\sigma_{\min}=5\ cm$. We use standard cross-entropy loss as the loss function. The batch size of input voxel grids is 4 and the network is trained for 30 epochs. The Adam optimizer with learning rate $0.0001$ is used for updating the weights of the networks. This will take about 12 hours for training on our dataset.

\textbf{Testing.} During the testing process, multi-view images are fed into the trained model of MF-PIFu to generate occupancy predictions for a volume. Then, the predicted 3D human bodies are extracted by an iso-surface through marching cubes from the volume. After voxelizing the predicted 3D model to low-resolution with $128^3$, the low-resolution voxel grid is fed into the trained model of VSR to refine the occupancy predictions of the volume. Through use of the march cubes again, the final 3D human body model is extracted from the iso-surface of the volume. Therefore, this process is an image-based coarse-to-fine 3D human body reconstruction method. We firstly obtain a coarse 3D reconstruction from multi-view image through learning the implicit function. Then, based on the coarse 3D prediction, the VSR can refine the coarse results through learning another implicit function. After the VSR, the false reconstructed parts can be removed and the details of the appearance can be preserved. 

\section{Experimental Results}
In this section some experiments are presented to evaluate our method. We firstly introduce the datasets and metrics for training and testing. Then, several previous methods are used for comparison on the quantitative and qualitative results. Finally, we discuss several factors which may affect the performance of our methods.
\subsection{Datasets and Metrics}
\begin{figure}
\centering
\subfigure[Multi-view images]{
    \begin{minipage}[t]{0.25\linewidth}
        \centering
        \includegraphics[width=1.1in]{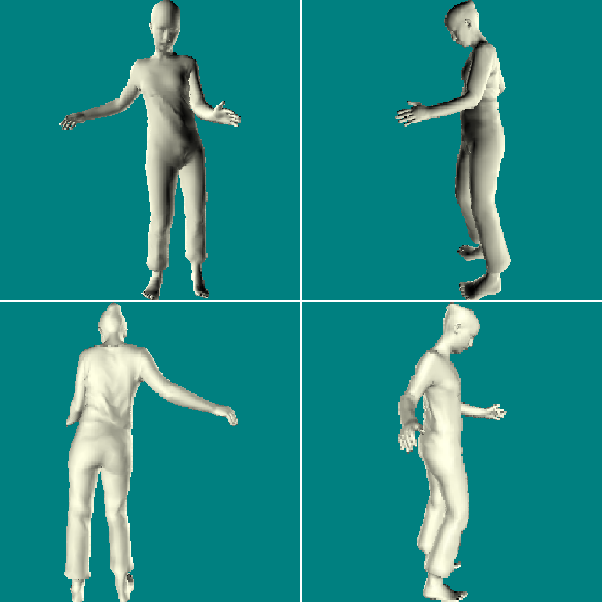}\\
        \includegraphics[width=1.1in]{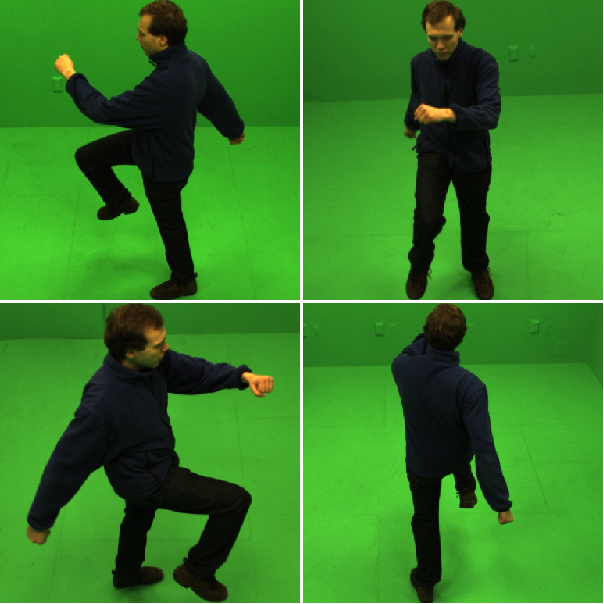}\\
    \end{minipage}%
}%
\subfigure[GT]{
    \begin{minipage}[t]{0.25\linewidth}
        \centering
        \includegraphics[width=1.1in]{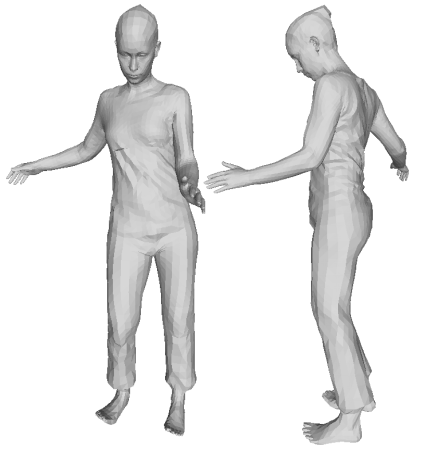}\\
        \includegraphics[width=1.1in]{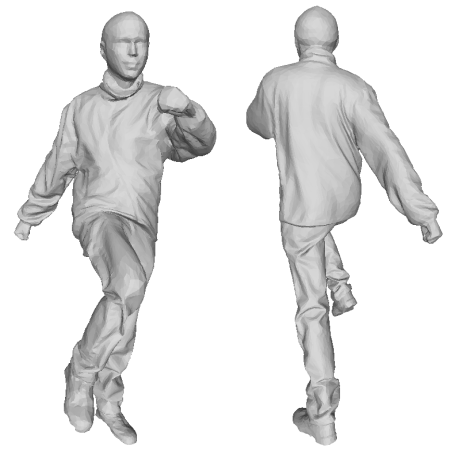}\\
    \end{minipage}%
}%
\subfigure[MF-PIFu]{
    \begin{minipage}[t]{0.25\linewidth}
        \centering
        \includegraphics[width=1.1in]{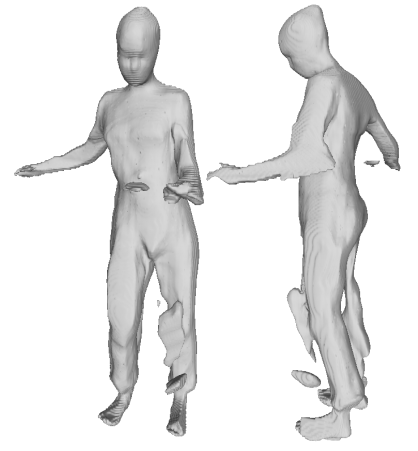}\\
        \includegraphics[width=1.05in]{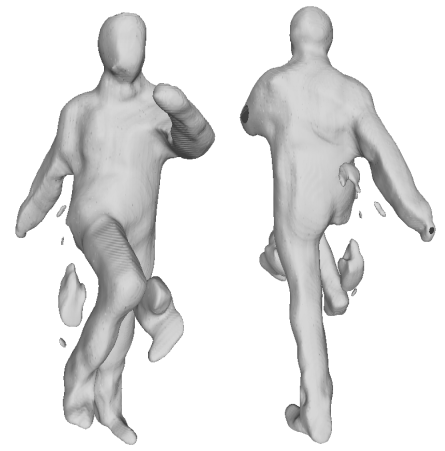}\\
        \vspace{0.03in}
    \end{minipage}%
}%
\subfigure[VSR]{
    \begin{minipage}[t]{0.25\linewidth}
        \centering
        \includegraphics[width=1.05in]{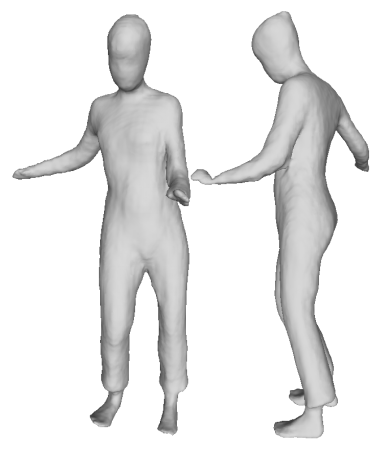}\\
        \includegraphics[width=1.1in]{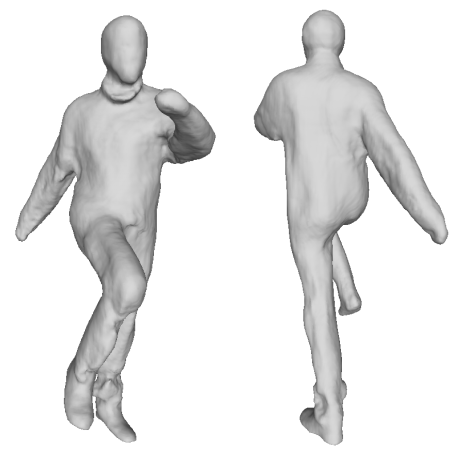}\\
    \end{minipage}%
}%
\centering
\caption{The 3D models from multi-view images and the 3D model after voxel super-resolution. From the left to right column: The original images (a), the ground truth of 3D model from two views (b), the estimated 3D models of MF-PIFu (c), and the final 3D model after VSR (d).}
\label{fig:result1}
\end{figure}

\textbf{Datasets.} To train and test our method, two datasets are used in the experiments: Articulated dataset~\cite{vlasic2008articulated} and CAPE dataset~\cite{ma2020learning}. Articulated dataset is captured by 8 cameras and it contains 10 indoor scenarios. Two male subjects have four scenarios, respectively, and one female subject performs two scenarios. For each scenario, RGB images, sillhouettes, camera parameters as well as 3D meshes are given. Totally, there are 2000 frames with eight-view images and 3D meshes. We split the dataset as 80\% frames (1600) for training and 20\% frames (400) for testing. The CAPE dataset is a 3D dynamic dataset of clothed humans generated by learning the clothing deformation from the SMPL body model. There are 15 generative clothed SMPL models with various poses. Since it has a large number of frames, we extract a small dataset from the original CAPE dataset. For each actions of each subject, we take the 80-$th$, 85-$th$, 90-$th$, 95-$th$, and 100-$th$ frames if the action has more than 100 frames. Totally, the small CAPE dataset has 2910 frames with 3D meshes. Since the dataset only provides 3D meshes, we render each mesh to four-view images from front, left, back and right side. Fig.~\ref{fig:result1} gives an example of four-view images and 3D mesh from the small CAPE dataset. We also split the dataset as 80\% for training and 20\% for testing in our experiments. 

\textbf{Metrics.} In order to evaluate our method quantitatively, we choose three metrices to measure the estimated 3D models: Euclidean distance from points on the estimated 3D models to surface of ground truth 3D mesh (P2S), Chamfer-$L_2$ and intersection over union between estimated 3D model and ground truth 3D model (IoU). For P2S and Chamfer-$L_2$, the lower value means the estimated 3D model is more accurate and complete. For IoU, the higher value means the estimated 3D model better match the ground truth. The detailed definition can be referred to~\cite{chibane20ifnet}.
\subsection{The results of the two steps}
In order to demonstrate the performance of MF-PIFu and VSR, we evaluate the results of the two parts on the two datasets. Fig.~\ref{fig:result1} gives the examples of the CAPE and Articulated dataset, respectively. The first row is an example from CAPE and the second row is an example from Articulated. The figure from left to right column shows (a) original multi-view images, (b) the ground truth of 3D mesh from two views, (c) the corresponding estimated 3D meshes by the MF-PIFu  and (d) the final results of VSR. We can see that the estimated 3D models by MF-PIFu are almost the same as the ground truth. However, there are still some false reconstruction and the details of appearance are not fully recovered, which can be seen from the two examples in Fig.~\ref{fig:result1} (c). For instance, the arms of the 3D model from the CAPE dataset are not fully reconstructed by MF-PIFu and there are some extra reconstructed parts around the legs of the 3D models from the Articulated dataset. From Fig.~\ref{fig:result1} (d), it shows that the results of VSR are refined. Those extra reconstruction in the estimated 3D models of MF-PIFu are removed and the details of the appearance are preserved, especially for arms of the 3D model for the CAPE example and the neck part of the 3D model for the Articulated example. Therefore, the refined models look more smooth and natural. This figure demonstrates that MF-PIFu can produce the coasrse 3D models from multi-view images and VSR can generate better results through refining the coarse 3D models.
\begin{table}
\caption{The quantitative results of the CAPE and Articulated datasets by the two steps of our method.}
\label{tab:tab1}
\begin{tabular}{l|l|l|l|l}
\hline
\multicolumn{2}{l|}{ \ } & P2S $\downarrow$  & Chamfer-$L_2$ $\downarrow$  & IoU $\uparrow$ \\
\hline
\multirow{2}{*}{CAPE} & MF-PIFu  & 0.9482 & 0.0196 & 0.7829 \\
\cline{2-5}
& VSR & \textbf{0.4954} & \textbf{0.0062} & \textbf{0.8440} \\
\hline
\multirow{2}{*}{Articulated} & MF-PIFu & 0.7332 & 0.0194 & 0.8484\\
\cline{2-5}
& VSR & \textbf{0.3754} & \textbf{0.0032} & \textbf{0.9051} \\
\hline
\end{tabular}
\end{table}
\begin{figure}[htbp]
\centering
\subfigure[Image]{
    \begin{minipage}[t]{0.16\linewidth}
        \centering
        \includegraphics[width=\textwidth]{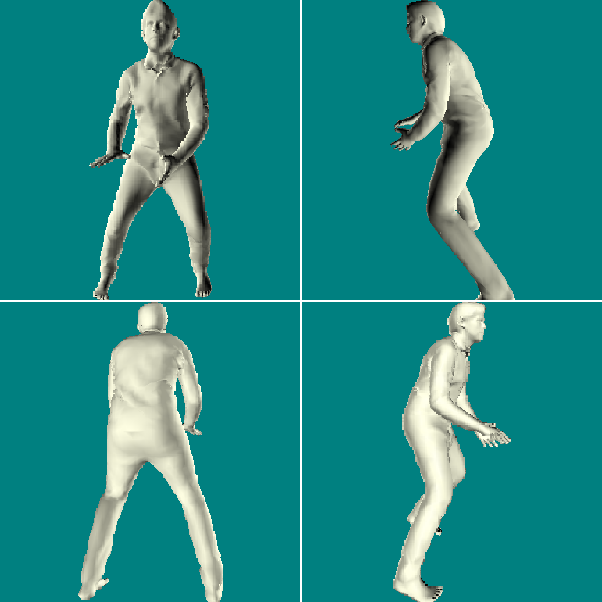}\\
        \includegraphics[width=\textwidth]{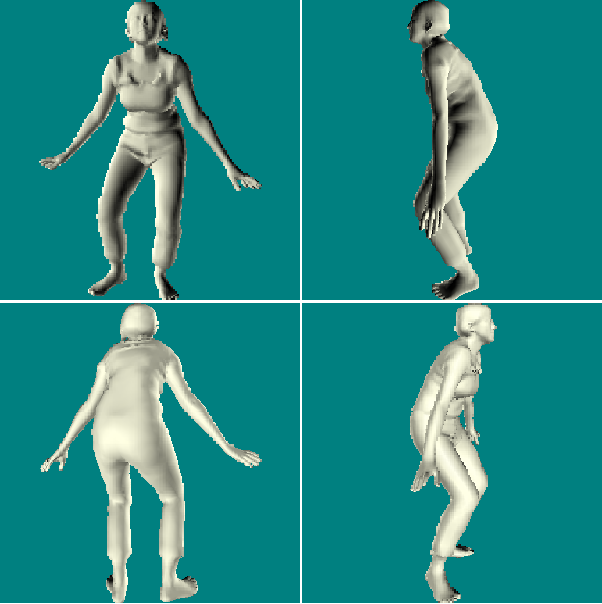}\\
        \includegraphics[width=\textwidth]{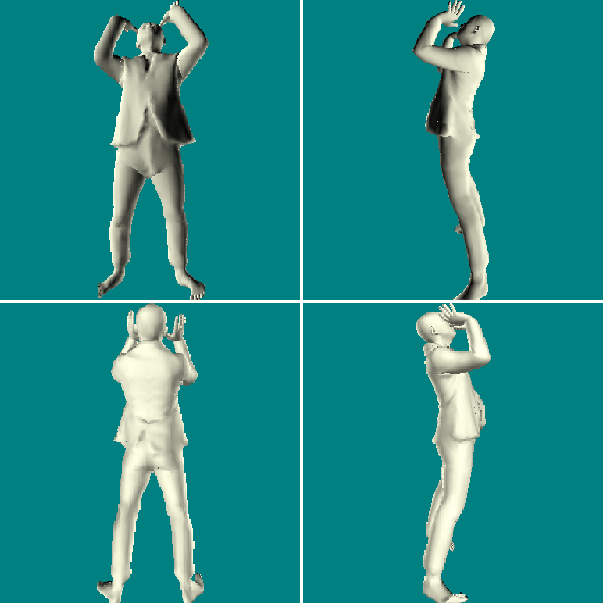}\\
        \vspace{0.05in}
    \end{minipage}%
}%
\subfigure[GT]{
    \begin{minipage}[t]{0.16\linewidth}
        \centering
        \includegraphics[width=0.7in]{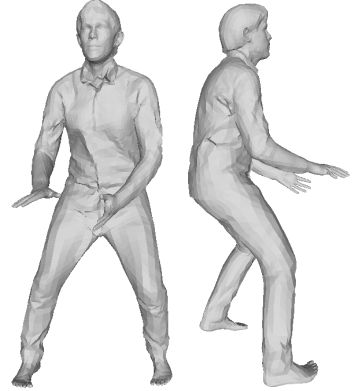}\\
        \includegraphics[width=0.8in]{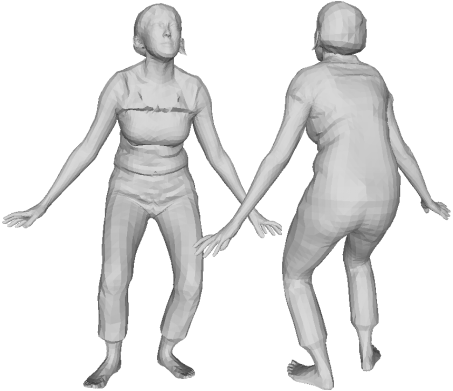}\\
        \vspace{0.05in}
        \includegraphics[width=0.61in]{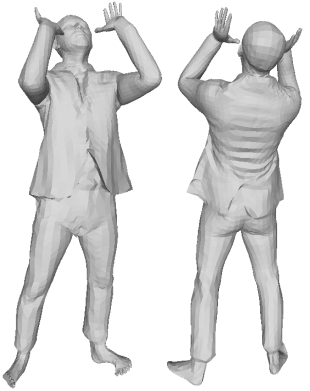}
        \vspace{0.035in}
    \end{minipage}%
}%
\subfigure[SPIN~\cite{Kolotouros_2019learning}]{
    \begin{minipage}[t]{0.16\linewidth}
        \centering
        \includegraphics[width=0.7in]{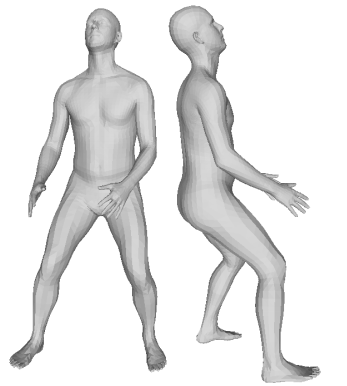}\\
        \includegraphics[width=0.72in]{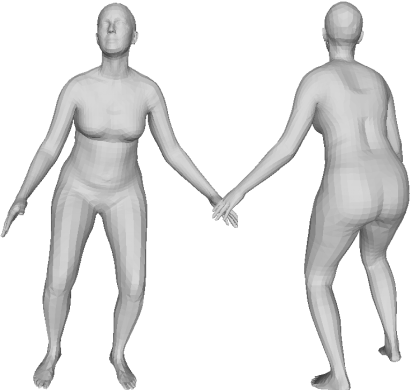}\\
        \vspace{0.04in}
        \includegraphics[width=0.6in]{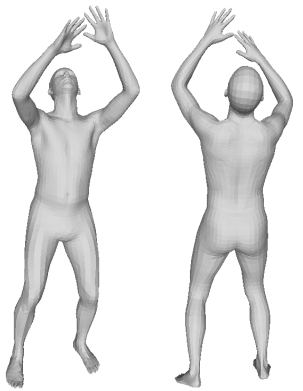}
        \vspace{0.05in}
    \end{minipage}%
}%
\subfigure[DeepHuman~\cite{zheng2019deephuman}]{
    \begin{minipage}[t]{0.16\linewidth}
        \centering
        \includegraphics[width=0.7in]{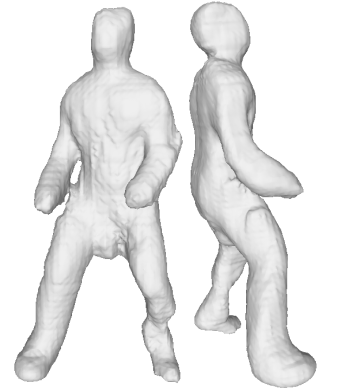}\\
        \includegraphics[width=0.7in]{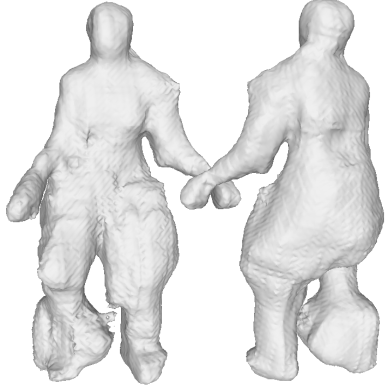}\\
        \vspace{0.02in}
        \includegraphics[width=0.62in]{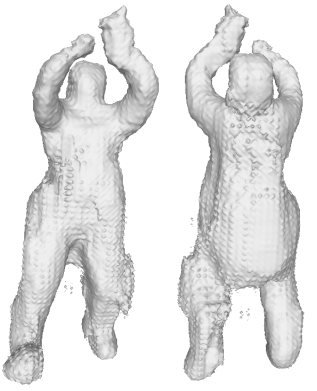}\\
        \vspace{0.055in}
    \end{minipage}%
}%
\subfigure[PIFu~\cite{saito2019pifu}]{
    \begin{minipage}[t]{0.16\linewidth}
        \centering
        \includegraphics[width=0.7in]{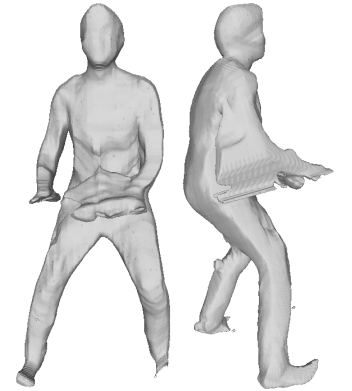}\\
        \includegraphics[width=0.8in]{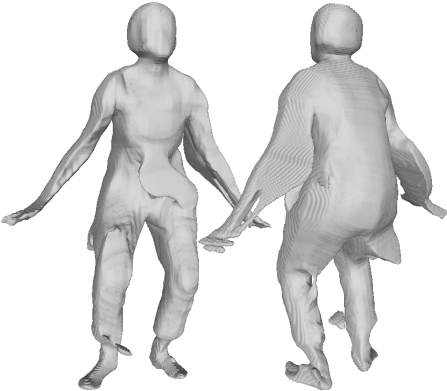}\\
        \vspace{0.02in}
        \includegraphics[width=0.61in]{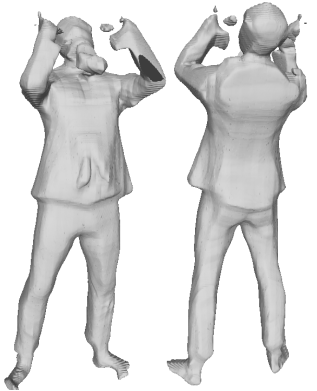}\\
        \vspace{0.07in}
    \end{minipage}%
}%
\subfigure[Our]{
    \begin{minipage}[t]{0.16\linewidth}
        \centering
        \includegraphics[width=0.7in]{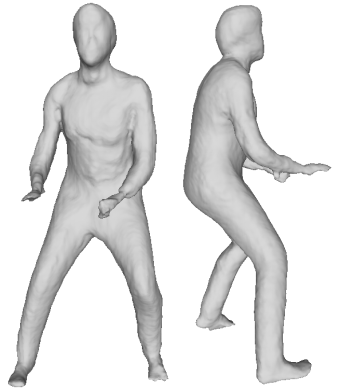}\\
        \includegraphics[width=0.8in]{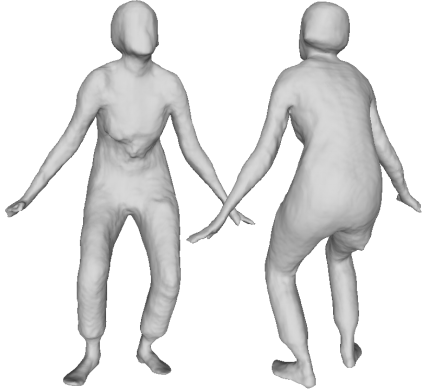}\\
        \vspace{0.02in}
        \includegraphics[width=0.61in]{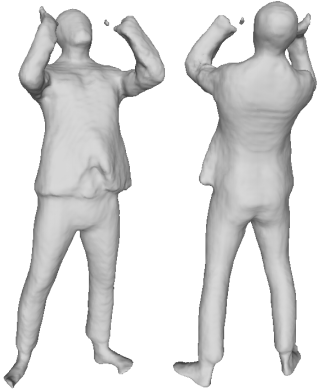}\\
        \vspace{0.04in}
    \end{minipage}%
}%
\centering
\caption{The comparison between our method and several previous methods on the CAPE dataset. Three examples are shown from top to down rows. The multi-view images, the ground truth of 3D models from two views, the estimated 3D models of SPIN~\cite{Kolotouros_2019learning}, DeepHuman~\cite{zheng2019deephuman}, PIFu~\cite{saito2019pifu} and our method are shown from the left to row column.}
\label{fig:result2}
\end{figure}

The quantitative results of the two steps on the two datasets are also shown in Table~\ref{tab:tab1}. The results of P2S, Chamfer-$L_2$ and IoU of the coarse 3D models by MF-PIFu and the refined 3D models of VSR are given in this table. We can see from the table that the P2S and Chamfer-$L_2$ of the VSR are smaller and the corresponding IoU is higher on both the two datasets. For the CAPE dataset, the P2S and Chamfer-$L_2$ after VSR decrease from 0.9428 \textit{cm} to 0.4954 \textit{cm} and from 0.0196 \textit{cm} to 0.0062 \textit{cm}, respectively. The IoU after VSR increases from 78.29\% to 84.40\%. For the Articulated dataset, the P2S and Chamfer-$L_2$ after VSR reduce from 0.7332 \textit{cm} to 0.3754 \textit{cm} and from 0.0194 \textit{cm} to 0.0032 \textit{cm}, respectively. The IoU after VSR increases from 84.29\% to 90.51\%. Therefore, the refined 3D models on the two datasets are more accurate and complete than the coarse 3D models. The VSR is useful to refine the models and obtains better 3D models. The conclusion of this table is consistent with Fig.~\ref{fig:result1}.

\subsection{Qualitative results}
\begin{figure}
\centering
\subfigure[Image]{
    \begin{minipage}[t]{0.16\linewidth}
        \centering
        \includegraphics[width=\textwidth]{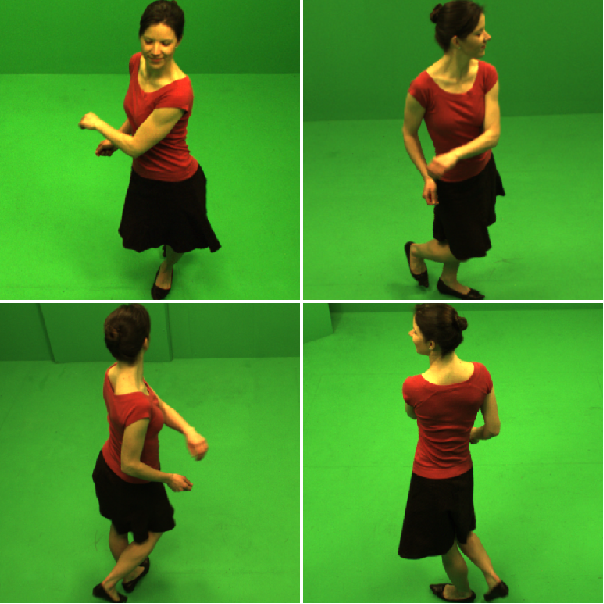}\\
        \includegraphics[width=\textwidth]{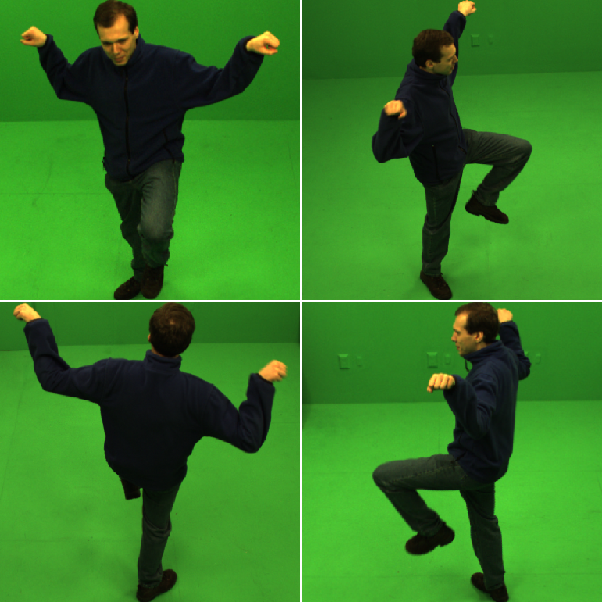}\\
        \includegraphics[width=\textwidth]{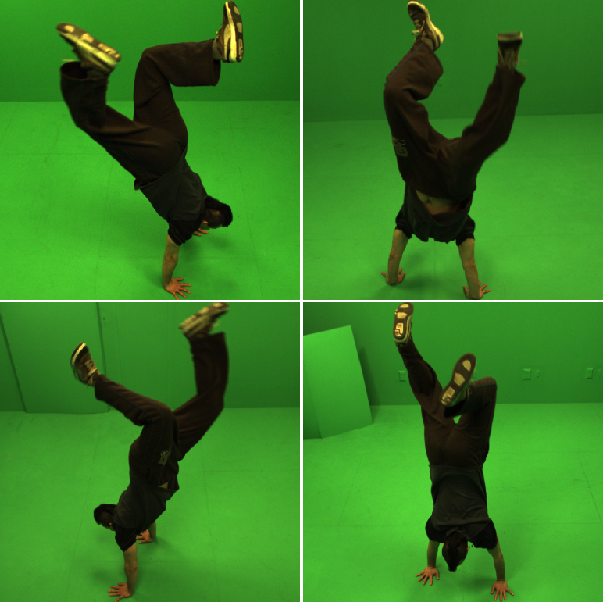}\\
        \vspace{0.05in}
    \end{minipage}%
}%
\subfigure[GT]{
    \begin{minipage}[t]{0.16\linewidth}
        \centering
        \includegraphics[width=0.6in]{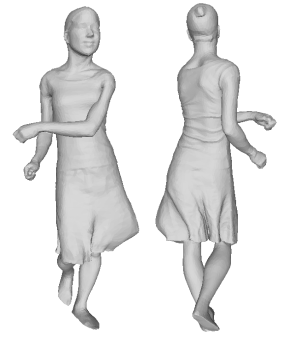}\\
        \vspace{0.07in}
        \includegraphics[width=0.8in]{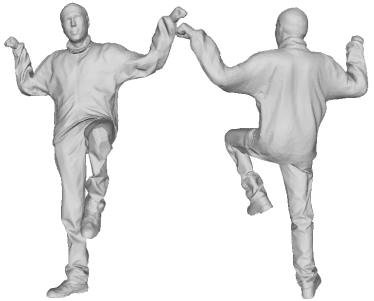}\\
        \vspace{0.05in}
        \includegraphics[width=0.7in]{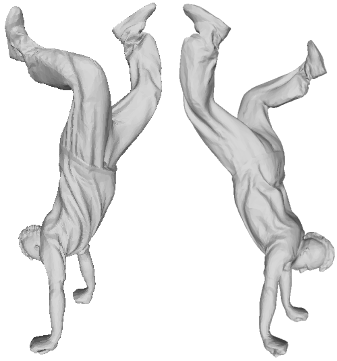}\\
        \vspace{0.05in}
    \end{minipage}%
}%
\subfigure[SPIN~\cite{Kolotouros_2019learning}]{
    \begin{minipage}[t]{0.16\linewidth}
        \centering
        \includegraphics[width=0.63in]{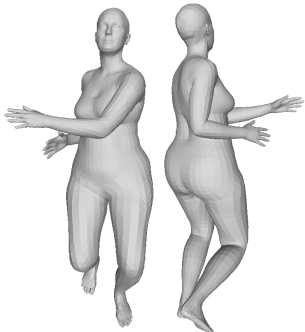}\\
        \vspace{0.07in}
        \includegraphics[width=0.85in]{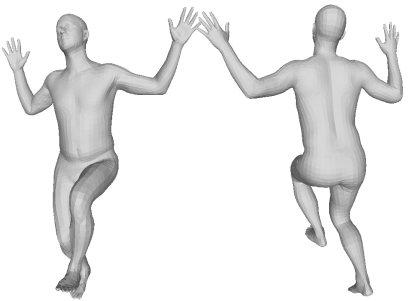}\\
        \vspace{0.05in}
        \includegraphics[width=0.7in]{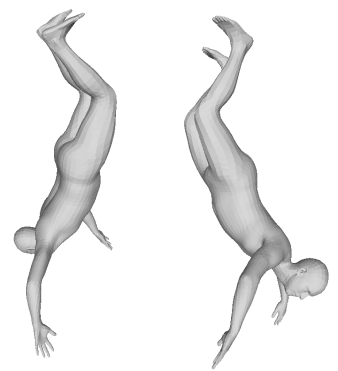}\\
        \vspace{0.046in}
    \end{minipage}%
}%
\subfigure[DeepHuman~\cite{zheng2019deephuman}]{
    \begin{minipage}[t]{0.16\linewidth}
        \centering
        \includegraphics[width=0.55in]{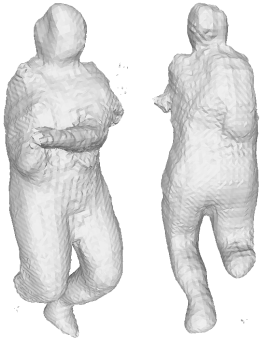}\\
        \vspace{0.07in}
        \includegraphics[width=0.8in]{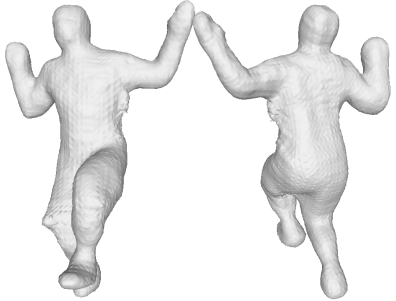}\\
        \vspace{0.06in}
        \includegraphics[width=0.75in]{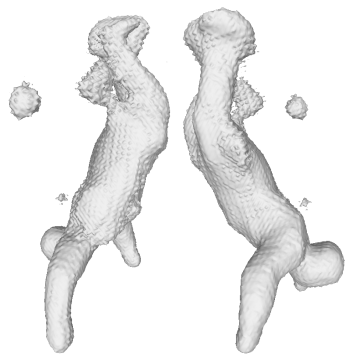}\\
        \vspace{0.05in}
    \end{minipage}%
}%
\subfigure[PIFu~\cite{saito2019pifu}]{
    \begin{minipage}[t]{0.16\linewidth}
        \centering
        \includegraphics[width=0.6in]{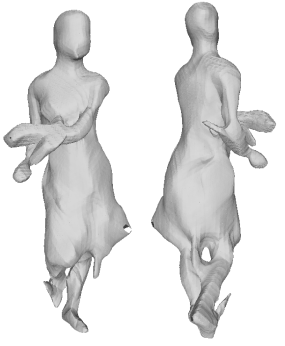}\\
        \vspace{0.07in}
        \includegraphics[width=0.8in]{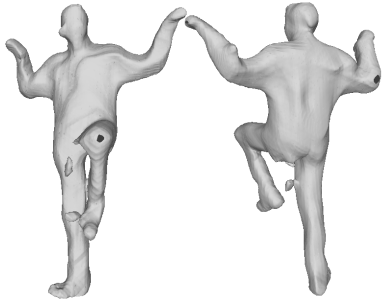}\\
        \vspace{0.05in}
        \includegraphics[width=0.75in]{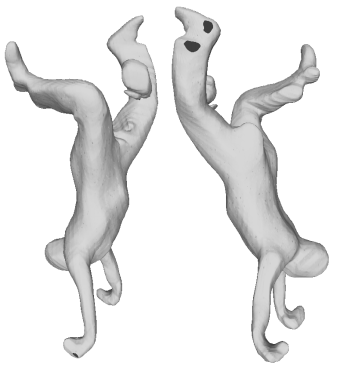}\\
        \vspace{0.035in}
    \end{minipage}%
}%
\subfigure[Our]{
    \begin{minipage}[t]{0.16\linewidth}
        \centering
        \includegraphics[width=0.6in]{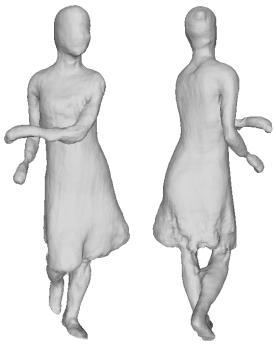}\\
        \vspace{0.07in}
        \includegraphics[width=0.8in]{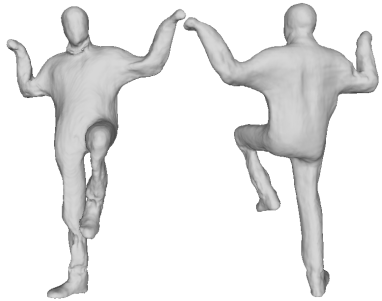}\\
        \vspace{0.05in}
        \includegraphics[width=0.72in]{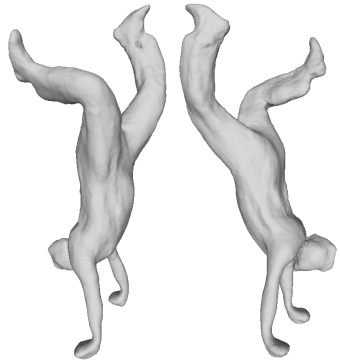}\\
        \vspace{0.05in}
    \end{minipage}%
}%
\centering
\caption{The comparison between our method and several previous methods on the Articulated dataset. Three examples are shown from top to down rows. The multi-view images, the ground truth of 3D models from two views, the estimated 3D models of SPIN~\cite{Kolotouros_2019learning}, DeepHuman~\cite{zheng2019deephuman}, PIFu~\cite{saito2019pifu} and our method are shown from the left to row column.}
\label{fig:result3}
\end{figure}
We qualitatively compare our method with several previous approaches for 3D human body reconstruction from images including SPIN~\cite{Kolotouros_2019learning}, DeepHuman~\cite{zheng2019deephuman} and PIFu~\cite{saito2019pifu}. For the SPIN and DeepHuman, we used the trained model provided by the authors to obtain the results. The two methods rely on the SMPL model~\cite{Loper_2015smpl} to reconstruct 3D human body from single images. For the PIFu, we trained and tested it on the same training dataset of Articulated and CPAE as our method from four-view images. The SPIN estimated the pose and shape parameters of SMPL model through collaborating regression and optimization. The estimated 3D models of SPIN are naked because the results of SPIN are the SMPL models parameterized by the estimated pose and shape parameters. The DeepHuman used encoder-decoder on the volume of deformed SMPL model and used normal image to refine the deformed SMPL model. This method can produce detailed SMPL model because the normal image could refine the appearance of SMPL model. In Fig.~\ref{fig:result2} and Fig.~\ref{fig:result3}, some examples from the two datasets and the results of SPIN~\cite{Kolotouros_2019learning}, DeepHuman~\cite{zheng2019deephuman}, PIFu~\cite{saito2019pifu} and our method are demonstrated, respectively. For each dataset, we give three examples which cover various poses and clothes to compare the performance of the methods. We can see that the estimated 3D models of SPIN and DeepHuman are not good enough but the results of PIFu and our method are better. Since the SPIN and DeepHuman rely on the SMPL model, they cannot handle the detailed appearance like clothes and wrinkles on the 3D models. Although DeepHuman attempts to recover the clothes on the 3D model, the results are not satisfying because the trained model of DeepHuman in the original paper is based on a different dataset. The results of PIFu are better than SPIN and DeepHuman because of learning an implicit representation, but there are some false parts in the results since the features in PIFu are at the same scales. By contrast, our method uses a coarse-to-fine manner to better reconstruct 3D human body models. The MF-PIFu estimates the coarse 3D models based on multi-scale features and implicit representation, and the VSR refines the coarse models to generate final results also based on multi-scale features and implicit representation. Our method can recover the 3D human body models from multi-view images with plausible pose and surface quality.
\begin{figure}[htbp]
\subfigure[SPIN~\cite{Kolotouros_2019learning}]{
    \begin{minipage}[t]{0.2\linewidth}
        \centering
        \includegraphics[width=0.5in]{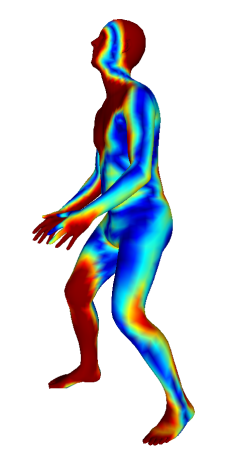}\\
        \includegraphics[width=0.5in]{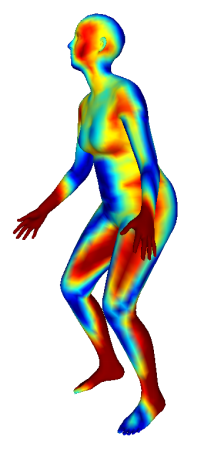}\\
        \includegraphics[width=0.5in]{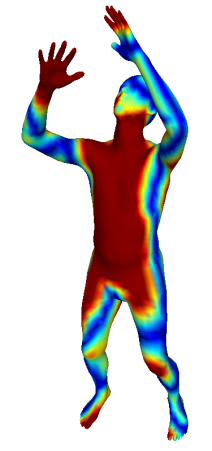}\\
        \includegraphics[width=0.5in]{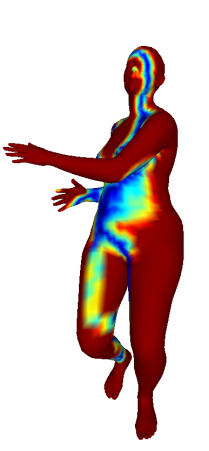}\\
        \includegraphics[width=0.5in]{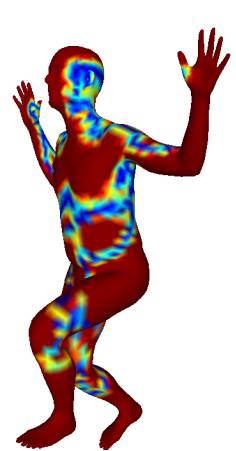}\\
        \includegraphics[width=0.5in]{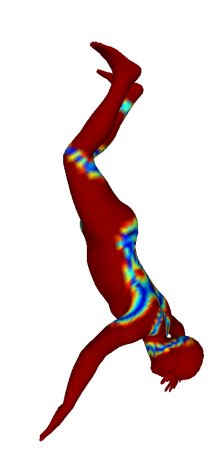}\\
        \vspace{0.02in}
    \end{minipage}%
}%
\subfigure[DeepHuman~\cite{zheng2019deephuman}]{
    \begin{minipage}[t]{0.2\linewidth}
        \centering
        \includegraphics[width=0.5in]{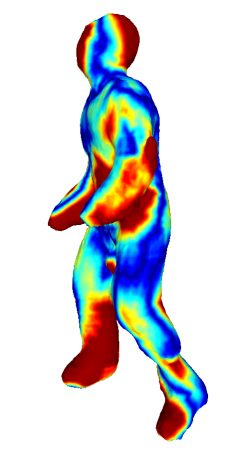}\\
        \includegraphics[width=0.5in]{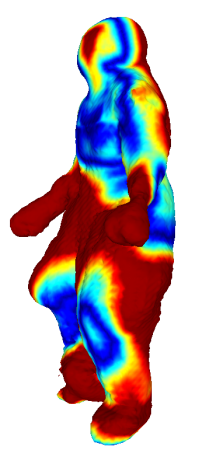}\\
        \includegraphics[width=0.5in]{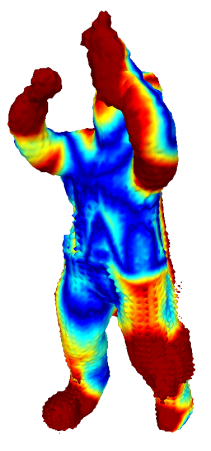}\\
        \includegraphics[width=0.5in]{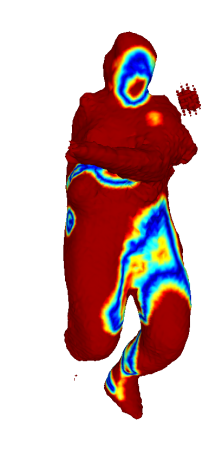}\\
        \includegraphics[width=0.5in]{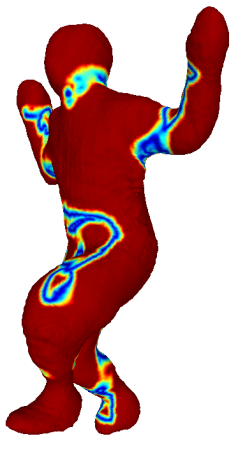}\\
        \includegraphics[width=0.5in]{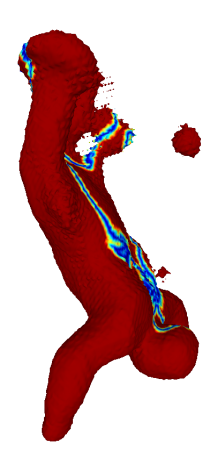}\\
        \vspace{0.035in}
    \end{minipage}%
}%
\subfigure[PIFu~\cite{saito2019pifu}]{
    \begin{minipage}[t]{0.2\linewidth}
        \centering
        \includegraphics[width=0.5in]{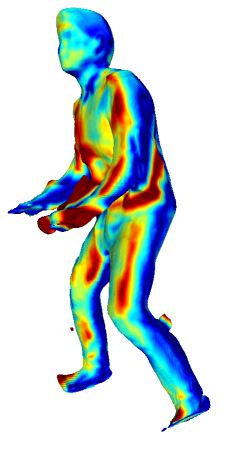}\\
        \includegraphics[width=0.5in]{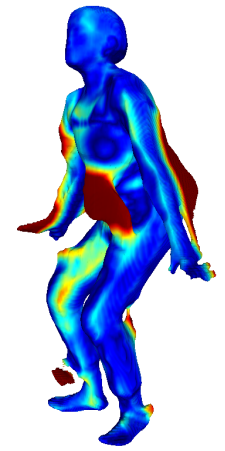}\\
        \includegraphics[width=0.5in]{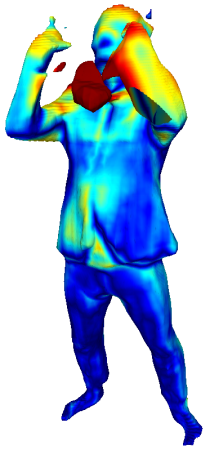}\\
        \includegraphics[width=0.5in]{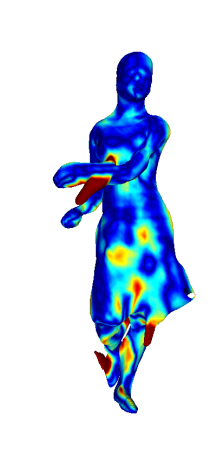}\\
        \includegraphics[width=0.5in]{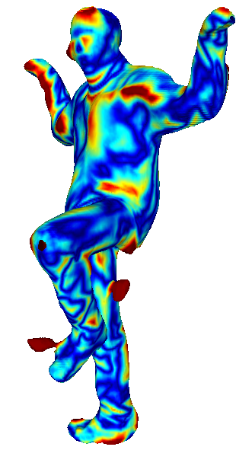}\\
        \includegraphics[width=0.5in]{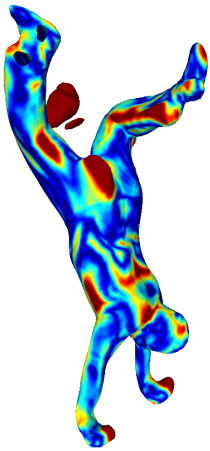}\\
        \vspace{0.029in}
    \end{minipage}%
}%
\subfigure[Our]{
    \begin{minipage}[t]{0.2\linewidth}
        \centering
        \includegraphics[width=0.5in]{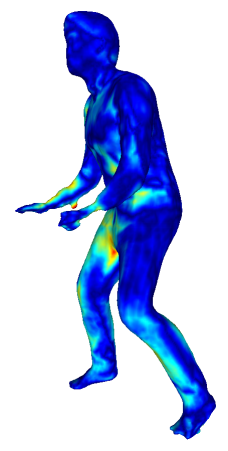}\\
        \includegraphics[width=0.5in]{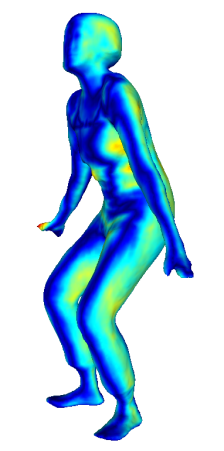}\\
        \includegraphics[width=0.5in]{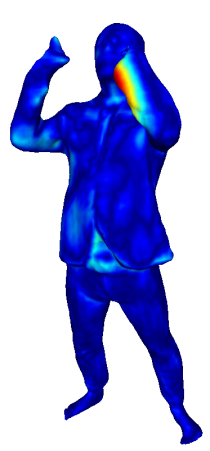}\\
        \includegraphics[width=0.5in]{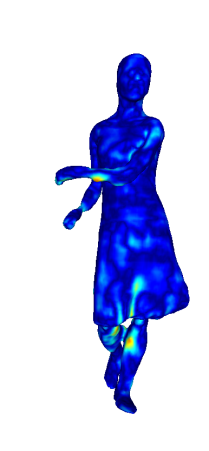}\\
        \includegraphics[width=0.5in]{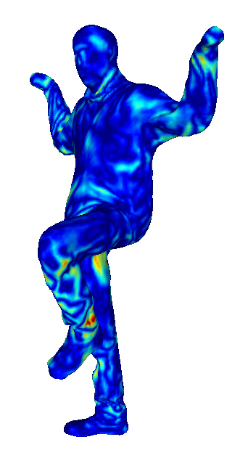}\\
        \includegraphics[width=0.5in]{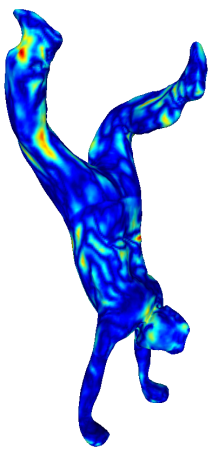}\\
    \end{minipage}%
}%
\caption{Visualization of the P2S between the estimated 3D models and the ground truth for different methods in Fig.~\ref{fig:result2} and Fig.~\ref{fig:result3}. The distance are represented by the heatmaps in Meshlab and mapped to the estimated 3D models.}
\label{fig:result_qm}
\end{figure}

In Fig.~\ref{fig:result_qm}, we visualize the P2S between the reconstructed 3D models in Fig.~\ref{fig:result2} and Fig.~\ref{fig:result3} by the different methods and the ground truth.  We use Meshlab to visualize the P2S to show the accuracy of the estimated 3D models by different methods. In Meshlab, the P2S is computed through the Hausdorff Distance. The distances are shown by the heatmaps and are mapped to the reconstructed 3D models. For every sample, the color range of different methods is based on the value of the P2S of our method. The red parts stand for high errors and the blue parts mean small distance. The figure clearly shows that the estimated 3D human bodies of our method have higher accuracy than the other three previous methods.
\subsection{Quantitative results}
In addition to the qualitative comparison, we also quantitatively compare to previous methods through computing the P2S, Chamfer-$L_2$ and IoU of results by different methods on the testing datasets of CAPE and Articulated. Table~\ref{tab:result1} and Table~\ref{tab:result2} demonstrate the mean values of the above metrics of different methods on the testing dataset of CAPE and Articulated, respectively. For the CAPE, the results of DeepHuman~\cite{zheng2019deephuman} are the worst because the CAPE is a synthetic dataset, but the trained model of DeepHuman is based on a real dataset. The SPIN~\cite{Kolotouros_2019learning} is better than DeepHuman, but it is still worse than PIFu~\cite{saito2019pifu} and our method because the estimated 3D models of SPIN are naked and the poses of the estimated 3D models might not be accurate. Comparing to SPIN and 
DeepHuman, the results of PIFu are better because PIFu uses four-view images and represents the 3D model through learning implicit function. Our method achieves the best performance among these methods because VSR can refine the coarse results of MF-PIFu. Both MF-PIFu and VSR in our method extract multi-scale features and learn the implicit function from multi-view images. The coarse-to-fine manner is an efficient way to obtain better models. The P2S and Chamfer-$L_2$ are the smallest in our method, which means that the results of our method are more accurate. The IoU of our method is the highest, which means that the estimated 3D models are more complete. 
For the Articulated dataset, Table~\ref{tab:result2} shows similar conclusion. The SPIN and DeepHuman achieve similar level on the real dataset and PIFu is better than the above two methods. However, our method also achieves the smallest P2S and Chamfer-$L_2$ and the highest IoU on the Articulated dataset. The two tables demonstrate that our method had good performance on both synthetic and real datasets.
\begin{table}
\caption{The quantitative results of SPIN~\cite{Kolotouros_2019learning}, DeepHuman~\cite{zheng2019deephuman}, PIFu~\cite{saito2019pifu} and our method on the testing dataset of the CAPE. Our method achieves better performance.}
\label{tab:result1}
\begin{tabular}{l|l|l|l}
\hline
Method & P2S $\downarrow$  & Chamfer-$L_2$ $\downarrow$  & IoU $\uparrow$ \\
\hline
SPIN~\cite{Kolotouros_2019learning}  & 2.2134 & 0.1271 & 0.4044 \\
DeepHuman~\cite{zheng2019deephuman}  & 3.4028 & 0.1850 & 0.3861 \\
PIFu~\cite{saito2019pifu}  & 1.0330 & 0.0212 & 0.7571 \\
Ours & \textbf{0.4954} & \textbf{0.0062} & \textbf{0.8440} \\
\hline
\end{tabular}
\end{table}
\begin{table}
\caption{The quantitative results of SPIN~\cite{Kolotouros_2019learning}, DeepHuman~\cite{zheng2019deephuman}, PIFu~\cite{saito2019pifu} and our method on the testing dataset of the Articulated. Our method achieves better performance.}
\label{tab:result2}
\begin{tabular}{l|l|l|l}
\hline
Method & P2S $\downarrow$  & Chamfer-$L_2$ $\downarrow$  & IoU $\uparrow$ \\
\hline
SPIN~\cite{Kolotouros_2019learning}  & 3.5206 & 0.2679 & 0.3506 \\
DeepHuman~\cite{zheng2019deephuman} & 3.9448 & 0.2675 & 0.3742 \\
PIFu~\cite{saito2019pifu}  & 0.8194 & 0.0210 & 0.8255 \\
Ours & \textbf{0.3754} & \textbf{0.0032} & \textbf{0.9051} \\
\hline
\end{tabular}
\end{table}

\begin{figure}
\subfigure[The P2S of the testing dataset of the CAPE for different methods.]{
    \begin{minipage}[t]{0.7\linewidth}
        \centering
        \includegraphics[width=\textwidth]{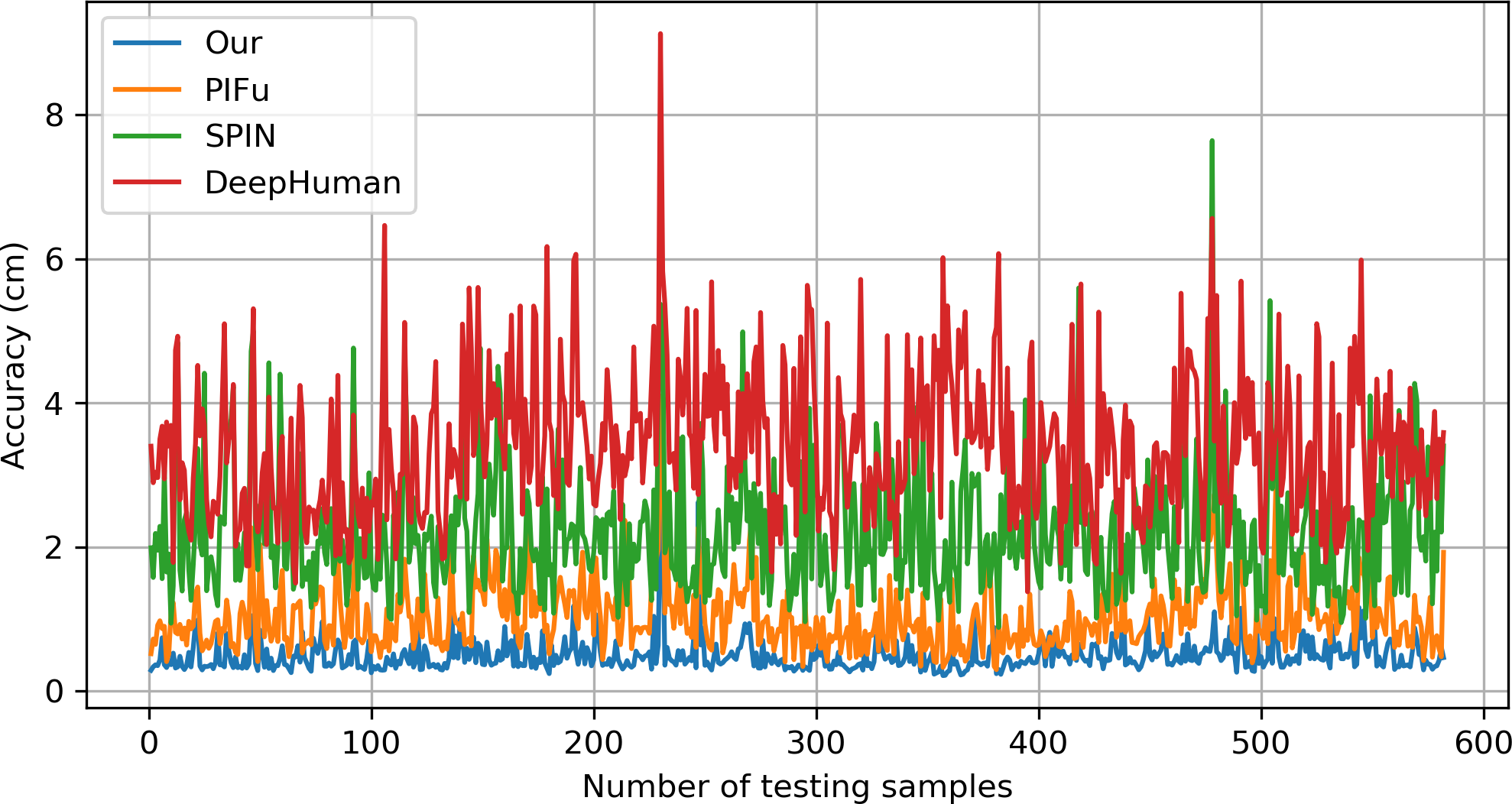}\\
    \end{minipage}%
}\\
\subfigure[The P2S of the testing dataset of the Articulated for different methods.]{
    \begin{minipage}[t]{0.7\linewidth}
        \centering
        \includegraphics[width=\textwidth]{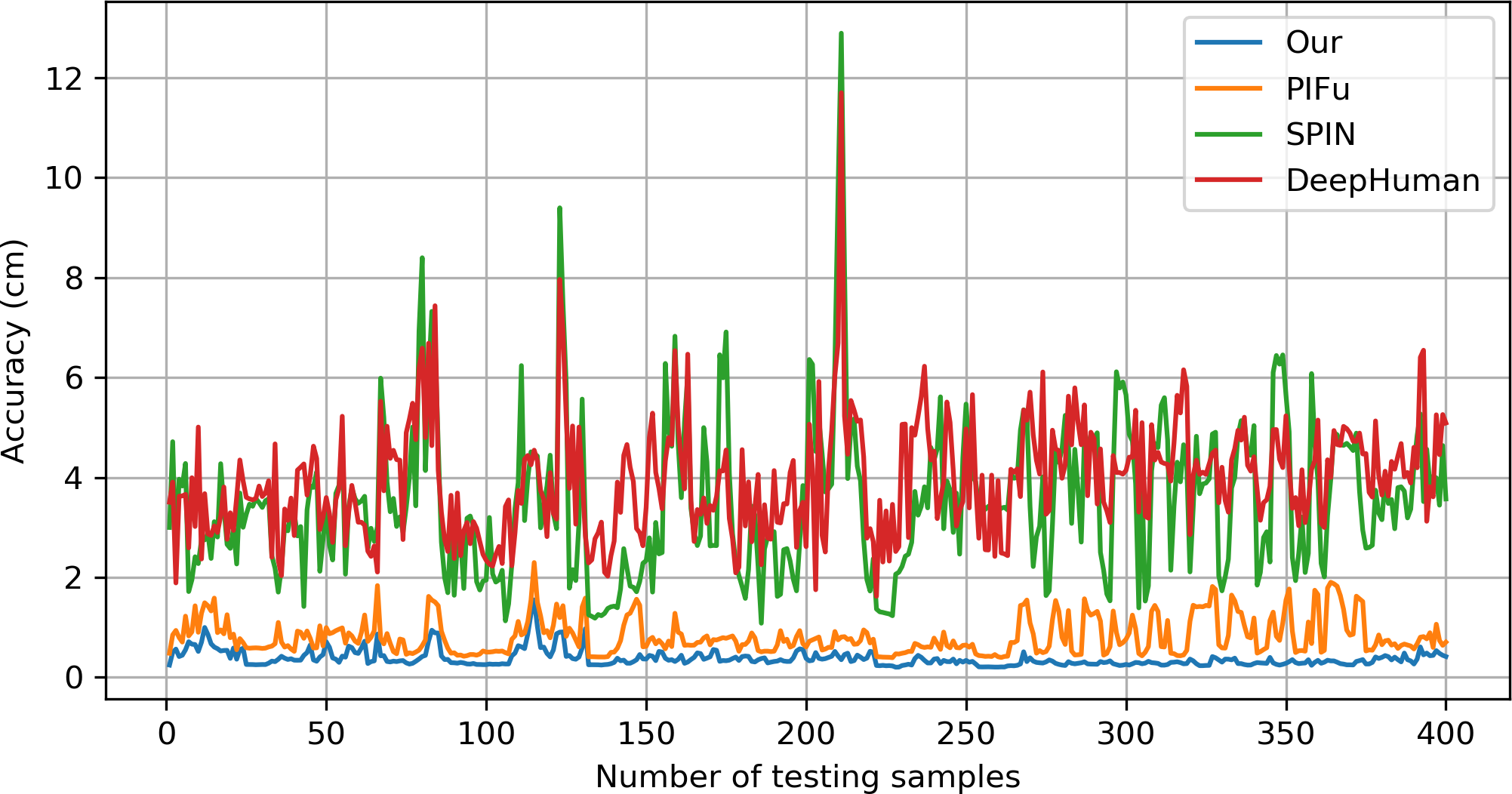}\\
    \end{minipage}%
}%
\caption{The P2S of each sample in the testing data of the two datasets for different methods. The $y$ axis stands for the accuracy of P2S. The $x$ axis is the number of samples in the testing data.}
\label{fig:eval}
\end{figure}
In order to clearly show the metric on the testing datasets, the P2S of each sample in the two testing data of the CAPE and Articulated dataset is shown in Fig.~\ref{fig:eval}. There are 582 samples in the testing dataset of CAPE and 400 samples in the testing dataset of Articulated, respectively. Our method (the blue line) has the lowest errors on the two datasets comparing to the other methods. Besides, for the testing samples, our method is more stable and robust because the blue lines do not have serious fluctuation.

\subsection{Discussion on the PIFu}
\begin{figure}
\includegraphics[width=0.7\linewidth]{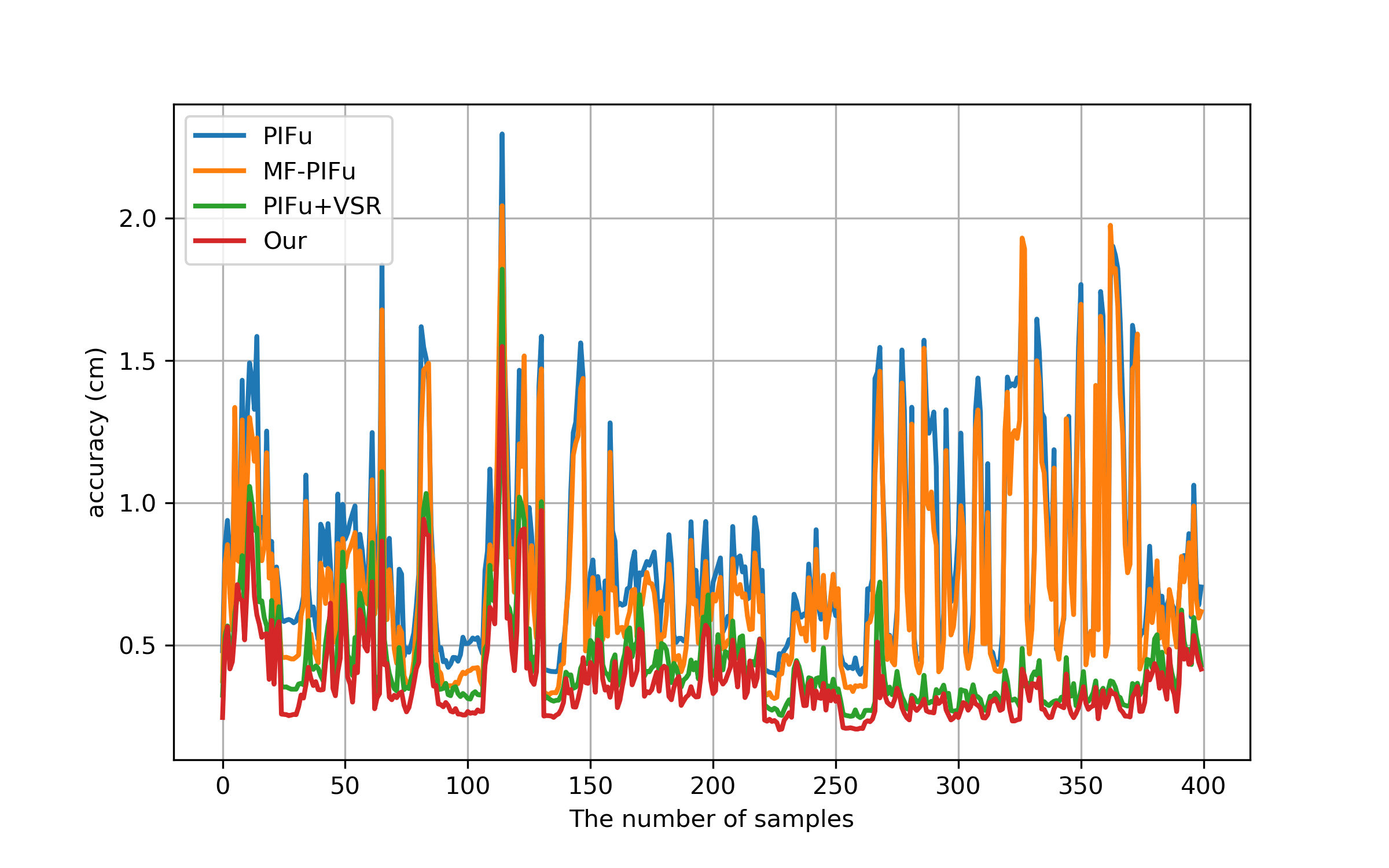}\\
\caption{The P2S of each sample in the testing data of the Articulated for PIFu, MF-PIFu, PIFu+VSR, and our method. The $y$ axis stands for the accuracy of P2S. The $x$ axis is the number of samples in the testing data.}
\label{fig:pifu}
\end{figure}

As shown above, PIFu~\cite{saito2019pifu} is a similar approach which also learns an implicit representation for 3D model from images. Therefore, we discuss more about the performance of PIFu in this section. The results of PIFu, MF-PIFu, PIFu+VSR and our method are evaluated to demonstrate the advantage of MF-PIFu and our method on the Articulated dataset. Table~\ref{tab:pifu} gives the quantitative results of PIFu, MF-PIFu, PIFu+VSR and our method on the testing dataset of the Articulated. PIFu+VSR means that PIFu is trained by the same Articulated dataset as MF-PIFu, and the testing results of PIFu is refined by the VSR which was trained by the low-resolution voxel grids obtained by MF-PIFu. This table shows that MF-PIFu achieves better results than PIFu and the VSR can refine the coasrse models obtained by PIFu and MF-PIFu. Our method combines the MF-PIFu and VSR, and thus, our method achieves the best performance on the dataset. Fig.~\ref{fig:pifu} gives the the P2S of the four cases on the testing dataset of the Articulated. We can see from the figure that the accuracy of our method on most samples is the highest. For the MF-PIFu, it has smaller P2S on the most samples than the original PIFu, which provides more reliable inputs for the voxel super-resolution. Therefore, our method combining MF-PIFu and VSR achieves the smallest P2S on most samples. This is consistent with  Table~\ref{tab:pifu}.

\begin{figure}[htbp]
\subfigure[GT]{
    \begin{minipage}[f]{0.14\linewidth}
        \centering
        \vspace{0.5in}
        \includegraphics[width=\textwidth]{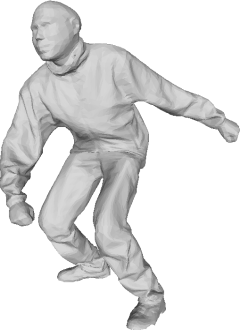}\\
        \vspace{0.96in}
        \includegraphics[width=\textwidth]{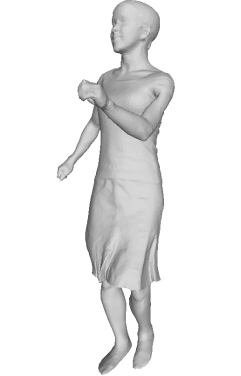}\\
        \vspace{0.5in}
    \end{minipage}%
}
\subfigure[PIFu]{
    \begin{minipage}[f]{0.14\linewidth}
        \centering
        \includegraphics[width=\textwidth]{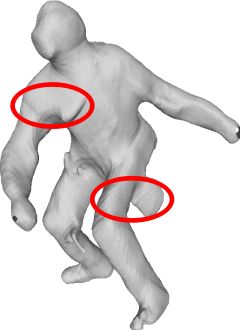}\\
        \includegraphics[width=\textwidth]{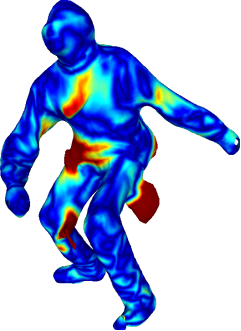}\\
        \includegraphics[width=\textwidth]{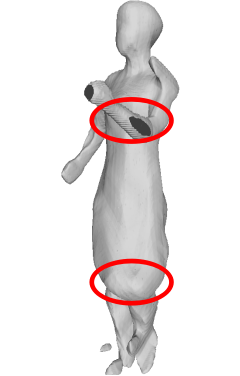}\\
        \includegraphics[width=\textwidth]{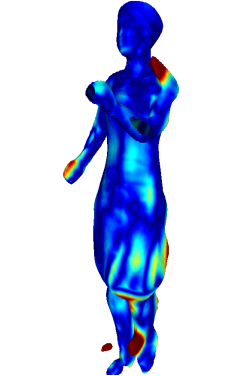}\\
    \end{minipage}%
}
\subfigure[MF-PIFu]{
    \begin{minipage}[f]{0.14\linewidth}
        \centering
        \includegraphics[width=\textwidth]{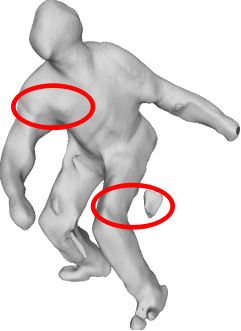}\\
        \includegraphics[width=\textwidth]{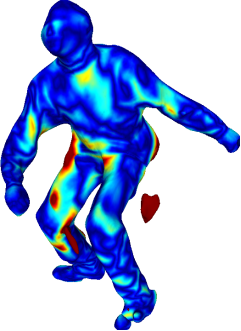}\\
        \includegraphics[width=\textwidth]{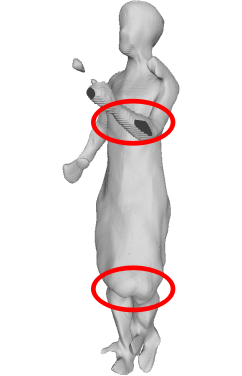}\\
        \includegraphics[width=\textwidth]{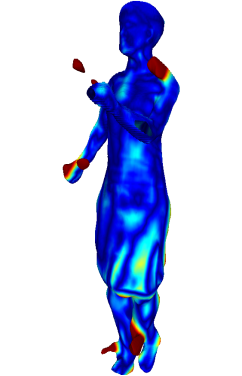}\\
    \end{minipage}%
}
\subfigure[PIFu+VSR]{
    \begin{minipage}[f]{0.14\linewidth}
        \centering
        \includegraphics[width=\textwidth]{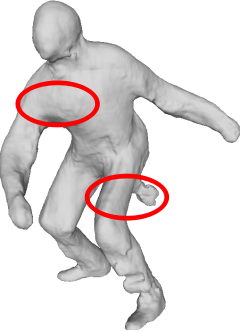}\\
        \includegraphics[width=\textwidth]{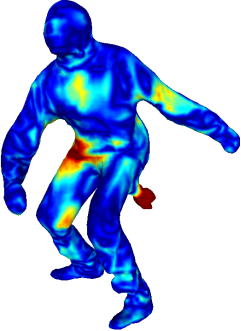}\\
        \includegraphics[width=\textwidth]{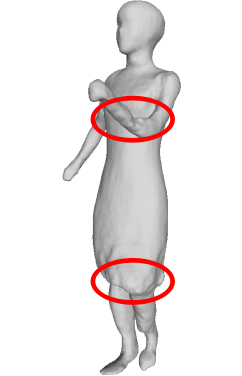}\\
        \includegraphics[width=\textwidth]{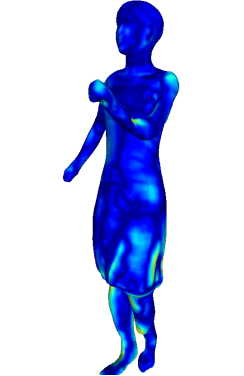}\\
    \end{minipage}%
}
\subfigure[Our]{
    \begin{minipage}[f]{0.14\linewidth}
        \centering
        \includegraphics[width=\textwidth]{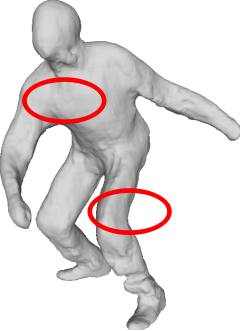}\\
        \includegraphics[width=\textwidth]{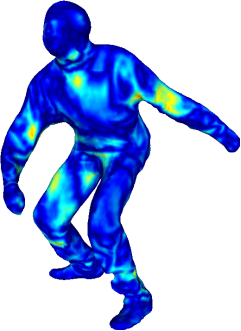}\\
        \includegraphics[width=\textwidth]{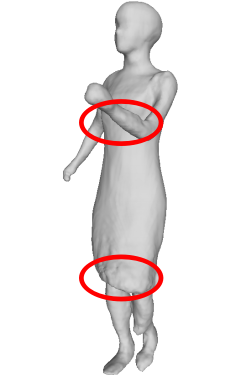}\\
        \includegraphics[width=\textwidth]{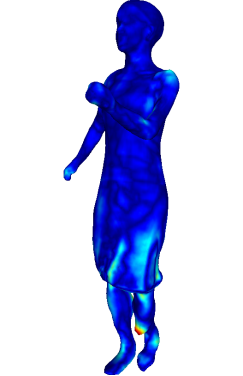}\\
    \end{minipage}%
}
\caption{The qualitative results of PIFu, MF-PIFu, PIFu+VSR, and our method on the Articulated dataset.}
\label{fig:pifu1}
\end{figure}

The qualitative examples from the Articulated dataset are shown in Fig.~\ref{fig:pifu1}. From the figure, it is clearly shown that the results of PIFu, MF-PIFu and PIFu+VSR have some false reconstruction, especially for the first example. The 3D models estimated by our method are the best because the false reconstruction is removed and the surface quality is improved by VSR, which can be demonstrated by the areas indicated by the red circles. The visualization of the errors on the 3D models is also given in the figure, which clearly shows that the 3D models of our method have the smallest distance to the ground truth among the four cases. 
\begin{table}
\caption{The qualitative results of PIFu, MF-PIFu, PIFu+VSR and our method.}
\label{tab:pifu}
\begin{tabular}{l|l|l|l}
\hline
View & P2S $\downarrow$  & Chamfer-$L_2$ $\downarrow$  & IoU $\uparrow$ \\
\hline
PIFu  & 0.8194 & 0.0210 & 0.8255 \\
MF-PIFu  & 0.7332 & 0.0194 & 0.8484\\
PIFu+VSR & 0.4322 & 0.0041 & 0.8865 \\
Our & \textbf{0.3754} & \textbf{0.0032} &  \textbf{0.9051}\\
\hline
\end{tabular}
\end{table}

\subsection{Spatial sampling}
\begin{figure}
\includegraphics[width=0.7\linewidth]{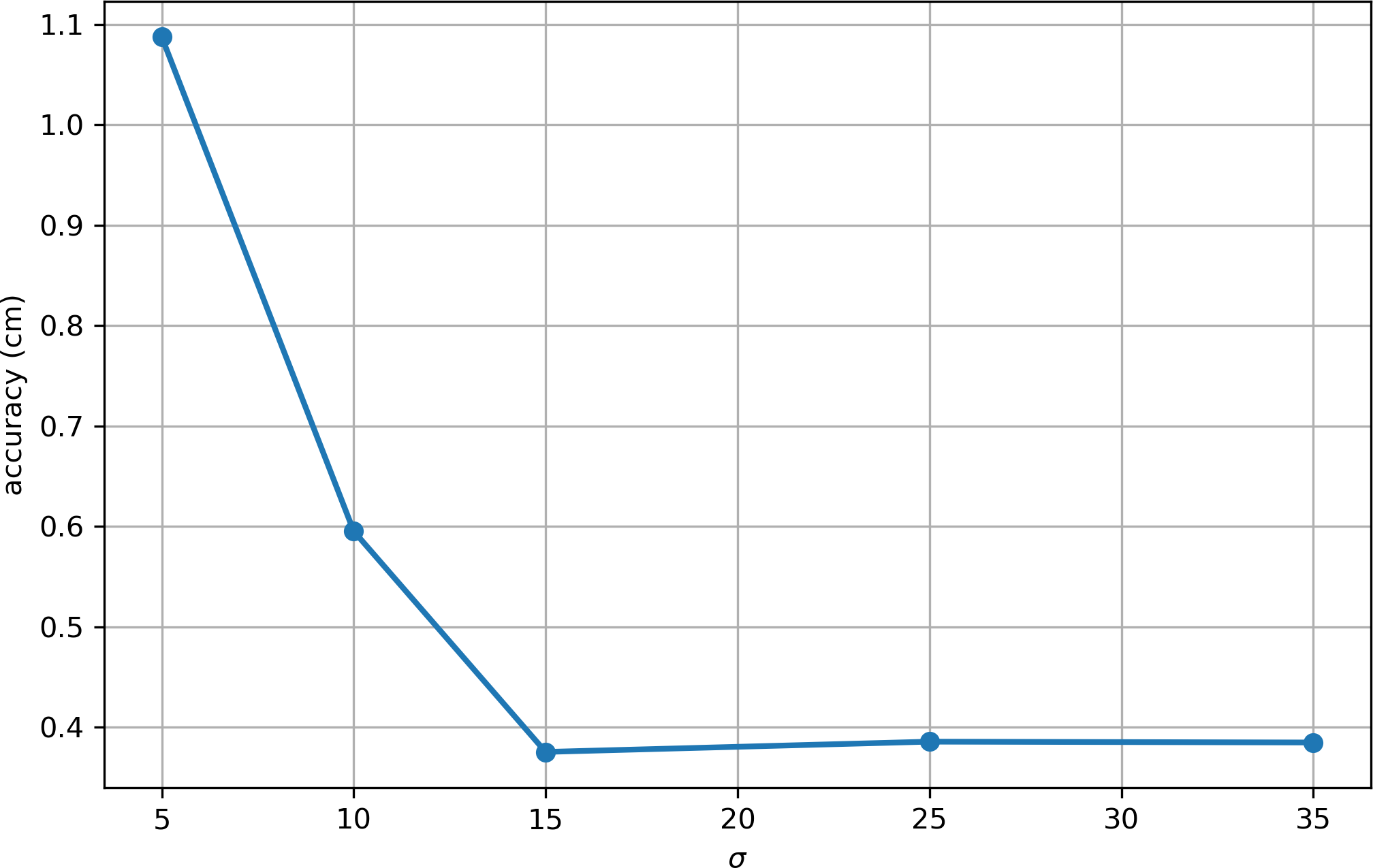}\\
\caption{The mean P2S on the testing dataset of the Articulated for different $\sigma_{\max}$. The $y$ axis stands for the mean P2S. The $x$ axis is the $\sigma_{\max}$.}
\label{fig:sigma}
\end{figure}

Spatial sampling is used in both MF-PIFu and VSR to generate the ground truth of the implicit value of spatial 3D points. It is an important factor in the sharpness of the final 3D model. In the two parts of our method, we use the same sampling strategy. Firstly, the points are uniformly sampled from the surface of the 3D model. Then, the random displacements with normal distribution $\mathcal{N}(0,\sigma)$ are added to the points. The $\sigma$ defines the distance of the points to the surface. The larger $\sigma$ makes the points further from the 3D mesh. For the MF-PIFu, we choose $\sigma=5\ cm$ for the random displacements because the paper of PIFu~\cite{saito2019pifu} has demonstrated that $\sigma=5\ cm$ can achieve the best performance for the 3D reconstruction from images. Here we evaluate the effects of $\sigma$ on the voxel super-resolution on the Articulated dataset. As shown in the implementation details, the 3D points are added random displacements with large $\sigma_{\max}$ and small $\sigma_{\min}$ during training the VSR. In order to discuss the effect of $\sigma_{\max}$ and $\sigma_{\min}$, we choose five pairs of $(\sigma_{\max},\sigma_{\min})$ and compare the corresponding performance under the five cases. Table~\ref{tab:sigma} shows the quantitative values of the P2S, Chamfer-$L_2$ and IoU for different $(\sigma_{\max},\sigma_{\min})$ on the testing dataset of the Articulated. Fig.~\ref{fig:sigma} shows the mean P2S of different $\sigma_{\max}$ for the testing dataset of the Articulated. The table and the figure demonstrate that the performance is almost the same for $(\sigma_{\max},\sigma_{\min})=(15,1.5),(25,2.5),(35,3.5)$. The P2S and IoU of the results for $(\sigma_{\max},\sigma_{\min})=(15,1.5)$ are the best, but it does not have too much difference with $(25,2.5)$ and $(35,3.5)$. This is the reason that we use $(\sigma_{\max},\sigma_{\min})=(15,1.5)$ in the quantitative and qualitative comparison to the previous methods.
\begin{table}
\caption{Quantitative results of different $(\sigma_{\max}, \sigma_{\min})$ on the Articulate dataset.}
\label{tab:sigma}
\begin{tabular}{l|l|l|l}
\hline
$(\sigma_{\max}, \sigma_{\min})$ (cm) & P2S $\downarrow$  & Chamfer-$L_2$ $\downarrow$  & IoU $\uparrow$ \\
\hline
(5,0.5) & 1.0874 & 0.1151 & 0.9006 \\
(10,1.0) & 0.5953 & 0.0110 & 0.8466 \\
(15,1.5) & \textbf{0.3754}  & 0.0032 & \textbf{0.9051} \\
(25,2.5) & 0.3856& 0.0030 & 0.8986 \\
(35,3.5) & 0.3848& \textbf{0.0029} & 0.8984 \\
\hline
\end{tabular}
\end{table}

\begin{figure}
\centering
\subfigure[GT]{
    \begin{minipage}[t]{0.09\linewidth}
        \centering
        \includegraphics[width=\textwidth]{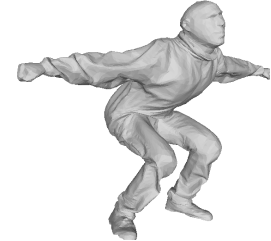}\\
        \includegraphics[width=\textwidth]{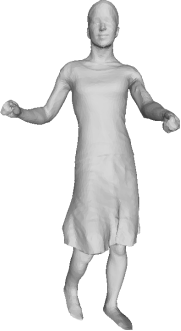}\\
    \end{minipage}%
}%
\subfigure[$\sigma_{\max}$=5]{
    \begin{minipage}[t]{0.18\linewidth}
        \centering
        \includegraphics[width=\textwidth]{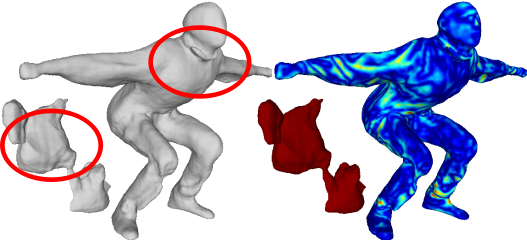}\\
        \includegraphics[width=\textwidth]{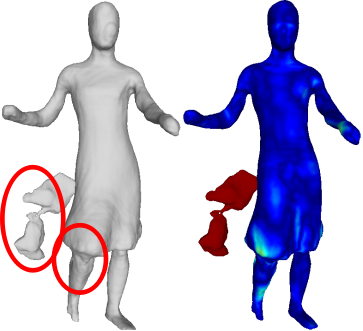}\\
    \end{minipage}%
}%
\subfigure[$\sigma_{\max}$=10]{
    \begin{minipage}[t]{0.18\linewidth}
        \centering
        \includegraphics[width=\textwidth]{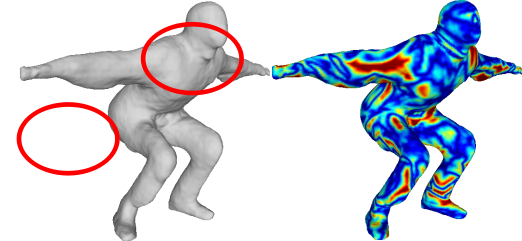}\\
        \includegraphics[width=\textwidth]{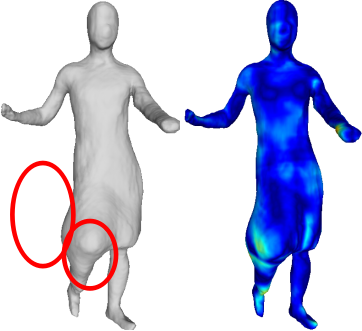}\\
    \end{minipage}%
}%
\subfigure[$\sigma_{\max}$=15]{
    \begin{minipage}[t]{0.18\linewidth}
        \centering
        \includegraphics[width=\textwidth]{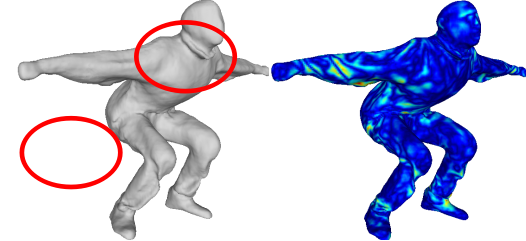}\\
        \includegraphics[width=\textwidth]{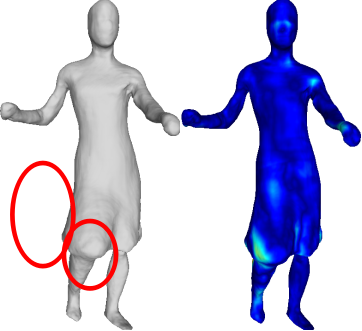}\\
    \end{minipage}%
}%
\subfigure[$\sigma_{\max}$=25]{
    \begin{minipage}[t]{0.18\linewidth}
        \centering
        \includegraphics[width=\textwidth]{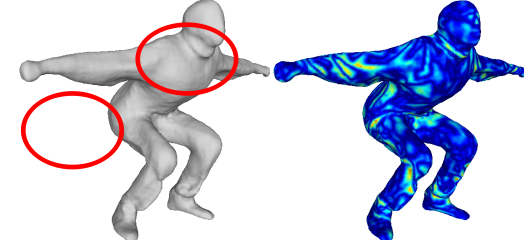}\\
        \includegraphics[width=\textwidth]{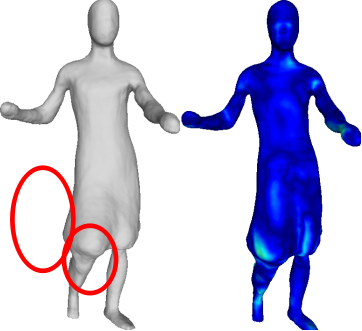}\\
    \end{minipage}%
}%
\subfigure[$\sigma_{\max}$=35]{
    \begin{minipage}[t]{0.18\linewidth}
        \centering
        \includegraphics[width=\textwidth]{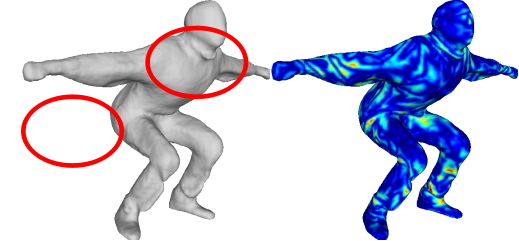}\\
        \includegraphics[width=\textwidth]{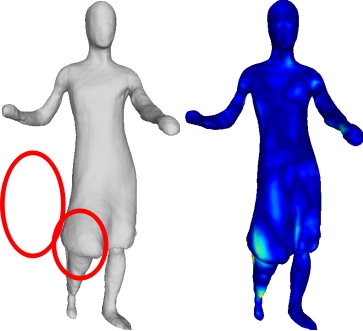}\\
    \end{minipage}%
}%
\centering
\caption{The comparison for differen $\sigma_{\max}$ on the Articulated dataset. From (a) to (f), two examples from the testing dataset are shown for $\sigma_{\max}=5,10,15,25,35$. For each $\sigma_{\max}$, the visualization of the error between the estimated result and the ground truth is given.}
\label{fig:sampling}
\end{figure}

Fig.~\ref{fig:sampling} shows two examples for different $\sigma$ from the Articulated dataset. We also give the visualization of the errors for the 3D models. From the figure, we can see that the estimated models of $\sigma_{\max}=5$ have extra unnecessary parts. The errors of $\sigma_{\max}=10$ are also relatively high from the visualization map, while the results of $\sigma_{\max}=15,25,35$ are almost the same level. However, as shown in the areas indicated by the red circles, the surface details of the estimated 3D models of $\sigma_{\max}=15$ are better preserved, especially for the neck part of the first example. Therefore, according to the above observation, the best choice for $(\sigma_{\max},\sigma_{\min})$ is $(15,1.5)$ for the Articulated dataset. It is also acceptable to use larger $(\sigma_{\max},\sigma_{\min})$, for instance, $(25,2.5)$ and $(35,3.5)$. However, this does not mean that $\sigma_{\max}$ can be too large because the results may not be good if $\sigma_{\min}$ is larger than $5\ cm$. The reasonable range for $(\sigma_{\max},\sigma_{\min})$ is $(15,1.5)\sim(35,3.5)$ according to the experiments.

\subsection{Voxel grid resolution}
\begin{figure}[htbp]
\subfigure[GT]{
    \begin{minipage}[t]{0.1\linewidth}
        \centering
        \includegraphics[width=\textwidth]{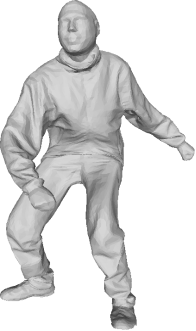}\\
        \includegraphics[width=\textwidth]{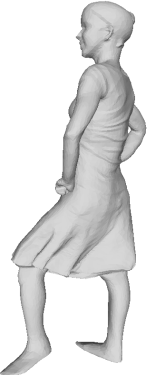}\\
        \includegraphics[width=\textwidth]{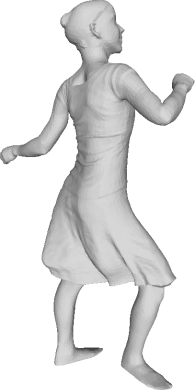}\\
        \vspace{0.02in}
    \end{minipage}%
}%
\subfigure[$32^3$]{
    \begin{minipage}[t]{0.1\linewidth}
        \centering
        \includegraphics[width=\textwidth]{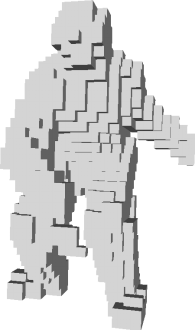}\\
        \includegraphics[width=\textwidth]{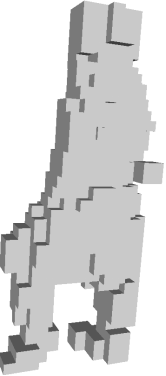}\\
        \includegraphics[width=\textwidth]{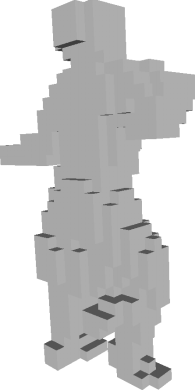}\\
    \end{minipage}%
}%
\subfigure[Results]{
    \begin{minipage}[t]{0.2\linewidth}
        \centering
        \includegraphics[width=\textwidth]{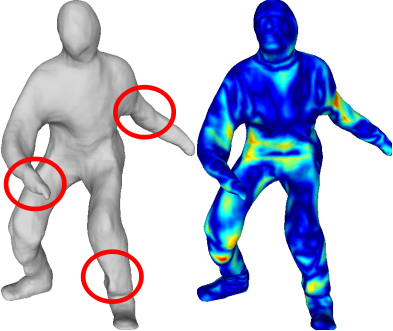}\\
        \includegraphics[width=\textwidth]{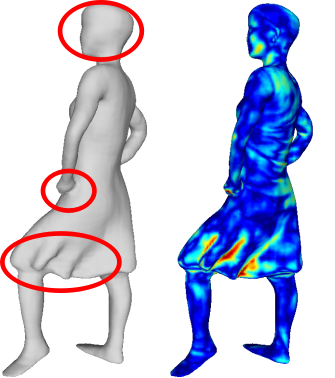}\\
        \includegraphics[width=\textwidth]{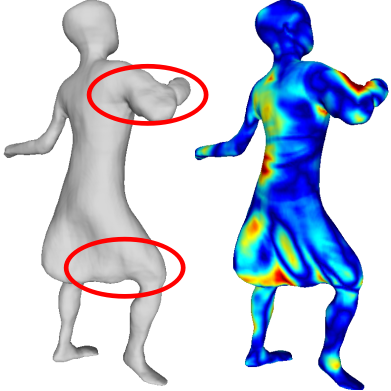}\\
        \vspace{0.02in}
    \end{minipage}%
}%
\subfigure[$128^3$]{
    \begin{minipage}[t]{0.1\linewidth}
        \centering
        \includegraphics[width=\textwidth]{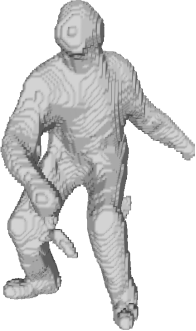}\\
        \includegraphics[width=\textwidth]{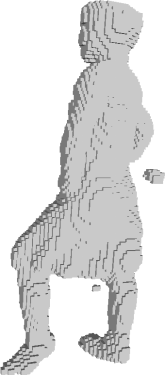}\\
        \includegraphics[width=\textwidth]{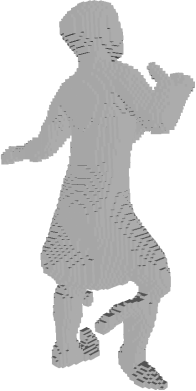}\\
    \end{minipage}%
}%
\subfigure[Results]{
    \begin{minipage}[t]{0.2\linewidth}
        \centering
        \includegraphics[width=\textwidth]{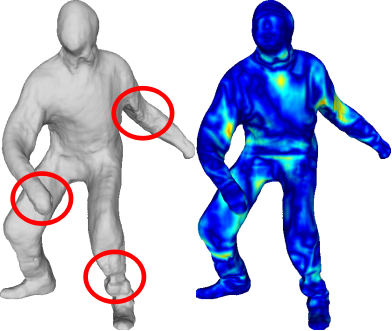}\\
        \includegraphics[width=\textwidth]{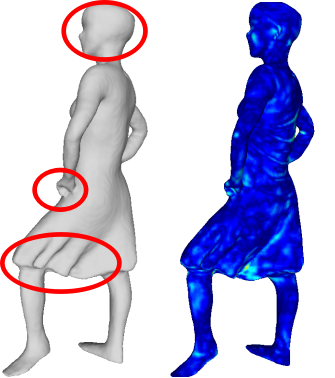}\\
        \includegraphics[width=\textwidth]{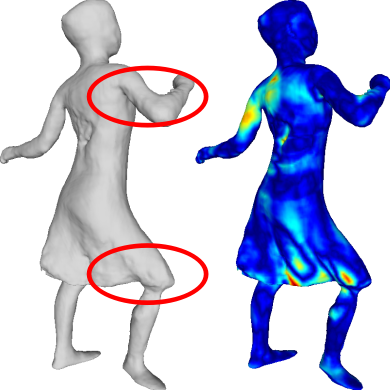}\\
        \vspace{0.02in}
    \end{minipage}%
}%
\caption{The comparison between $32^3$ and $128^3$ resolution on the Articulated dataset. (a) is the ground truth of 3D models; (b) is the voxel grids with $32^3$; (c) is the results of super resolution trained by $32^3$ voxel grids; (d) is the voxel grids with $128^3$; (e) is the results of super resolution trained by $128^3$ voxel grids.}
\label{fig:voxel}
\end{figure}

The resolution of input voxel grids for VSR will also affects the refinement of VSR to generate 3D models. In order to demonstrate the effects, we compare the results of VSR with the input resolution of $32^3$ and $128^3$ for the Articulated dataset. The voxel grids with different resolutions are generated from the estimated 3D models of MF-PIFu. Using the VSR which is trained by voxel grids with $128^3$, the final results are generated from voxel grids with $32^3$ and $128^3$, respectively. Table~\ref{tab:voxel} shows the P2S, Chamfer-$L_2$ and IoU of the results on the testing dataset of the Articulated for the input low-resolution voxel grids with $32^3$ and $128^3$ resolution. We can see that the quantitative values of the results for $128^3$ resolution are better than $32^3$. It is reasonable because higher resolution can provide more details for the voxel super-resolution. Fig.~\ref{fig:voxel} shows some examples of the $32^3$ and $128^3$ resolution. The 3D models after voxel super-resolution and the corresponding visualization of errors are shown in the figure. It also demonstrates that the results of VSR with $128^3$ resolution voxel grids has better details on the shape, especially for those areas indicated by the red circles. Therefore, the resolution of input voxel grid for voxel super-resolution should be as high as possible. In our observation, the resolution $128^3$ is reasonable to obtain good 3D model estimation considering the limitation of memory footprint.
\begin{table}
\caption{Quantitative results of $32^3$ and $128^3$ resolutions on the Articulate dataset.}
\label{tab:voxel}
\begin{tabular}{l|l|l|l}
\hline
voxel res. & P2S $\downarrow$  & Chamfer-$L_2$ $\downarrow$  & IoU $\uparrow$ \\
\hline
Ours($32^3$) & 1.9322 & 0.1626 & 0.6902 \\
Ours($128^3$) & \textbf{0.3754} & \textbf{0.0032} & \textbf{0.9051} \\
\hline
\end{tabular}
\end{table}
\subsection{The Number of images}
\begin{figure}
\includegraphics[width=0.6\linewidth]{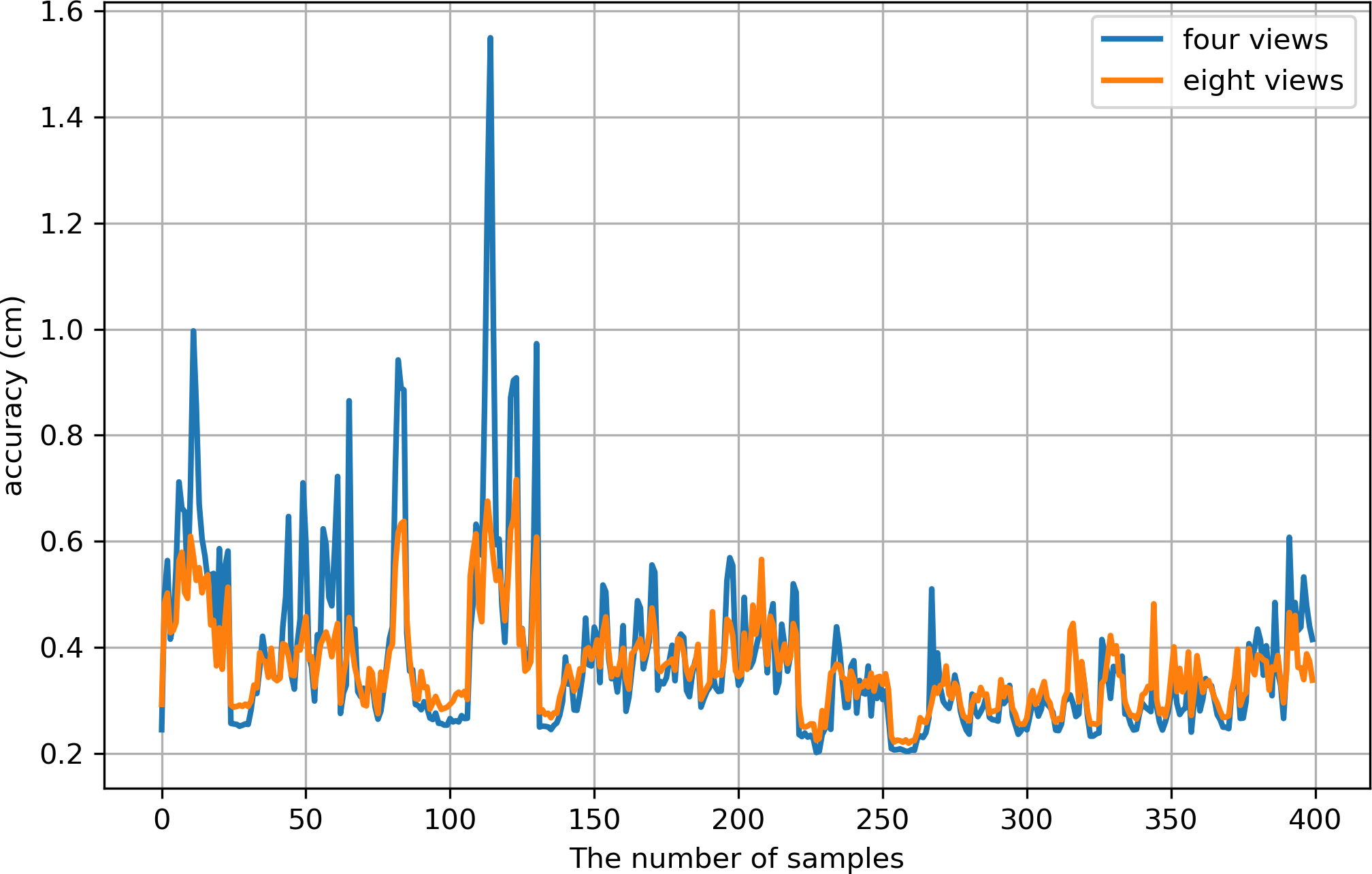}
\caption{The P2S of each sample in the testing data of the Articulated for four-view and eight-view images. The $y$ axis stands for the accuracy of P2S. The $x$ axis is the number of samples in the testing data.}
\label{fig:num_view}
\end{figure}

Since we estimate 3D human body from multi-view images, the effect of the number of views on the final estimation also needs to be discussed. We evaluate the performance of our method for four images and eight images on the Articulated dataset. Note that the MF-PIFu is trained by the four-view images and eight-view images, respectively. For the VSR, it is only trained by the voxel grids with $128^3$ resolution generated by the four-view images. Table~\ref{tab:num_view} shows the quantitative results on the Articulated dataset when the four-view and eight-view images are used. Fig.~\ref{fig:num_view} is the P2S of each sample in the testing dataset of Articulated for the four-view and eight-view cases. We can see that the results of eight-view case are a little better than the four-view case. Since eight-view images could provide more information for the MF-PIFu than the four-view images, the coarse 3D models obtained by MF-PIFu are more accurate, which ensures the coarse 3D models can provide more information for VSR to obtain better refined 3D models. During the voxel super-resolution, the training on the 3D space can help to reduce the ambiguity of four-view and eight-view cases. The final estimation does not have too much difference in the two cases. 

\begin{figure}
\subfigure[GT]{
    \begin{minipage}[t]{0.16\linewidth}
        \centering
        \includegraphics[width=\textwidth]{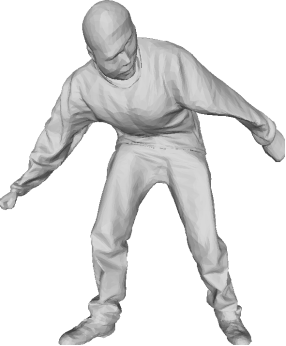}\\
        \includegraphics[width=\textwidth]{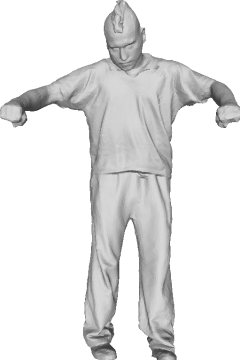}\\
    \end{minipage}%
}%
\subfigure[4-view 1]{
    \begin{minipage}[t]{0.16\linewidth}
        \centering
        \includegraphics[width=\textwidth]{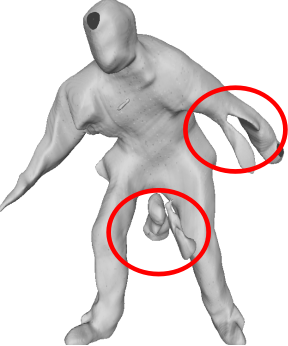}\\
        \includegraphics[width=\textwidth]{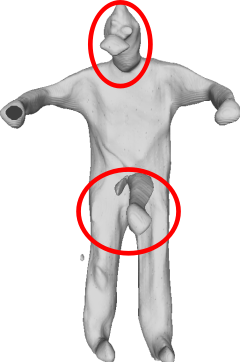}\\
    \end{minipage}%
}%
\subfigure[4-view 2]{
    \begin{minipage}[t]{0.16\linewidth}
        \centering
        \includegraphics[width=\textwidth]{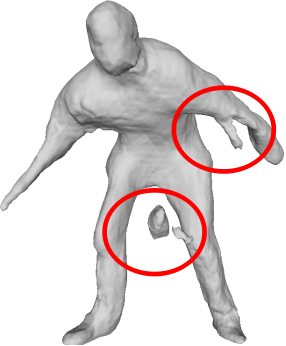}\\
        \includegraphics[width=\textwidth]{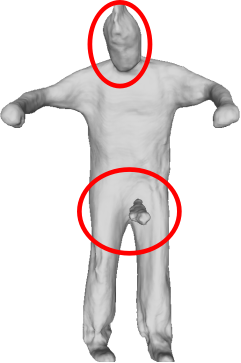}\\
    \end{minipage}%
}%
\subfigure[8-view 1]{
    \begin{minipage}[t]{0.16\linewidth}
        \centering
        \includegraphics[width=\textwidth]{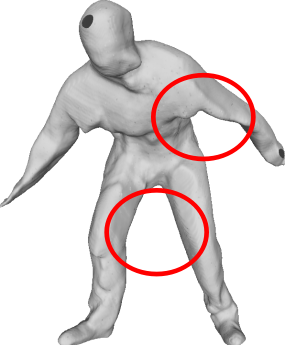}\\
        \includegraphics[width=\textwidth]{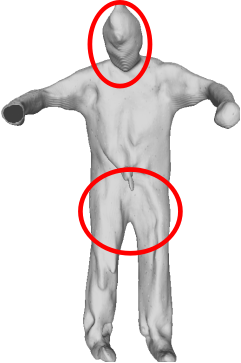}\\
    \end{minipage}%
}%
\subfigure[8-view 2]{
    \begin{minipage}[t]{0.16\linewidth}
        \centering
        \includegraphics[width=\textwidth]{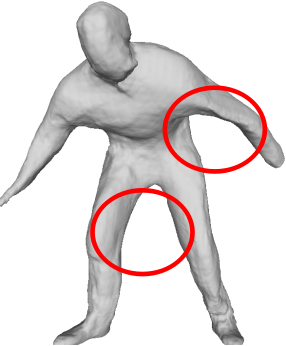}\\
        \includegraphics[width=\textwidth]{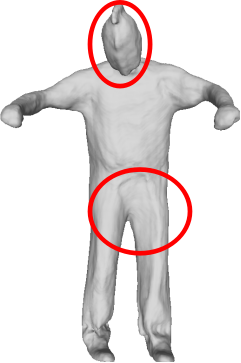}\\
    \end{minipage}%
}%
\caption{The results of four-view and eight-view images on the Articulated dataset. From left ot right columns: ground truth, the results of MF-PIFu of four-view images, the final results of four-view images, the results of MF-PIFu of eight-view images, and the final results of eight-view images.}
\label{fig:num_view1}
\end{figure}

Two examples from the Articulated dataset are shown in Fig.~\ref{fig:num_view1} for the four-view and eight-view images. The figure gives the results of MF-PIFu (b), the results of VSR (c) for the four-view images and the results of MF-PIFu (d), the results of VSR (e) for the eight-view images. We can see that there exists some error reconstruction on the 3D models of MF-PIFu for the four views, especially for the areas indicated by the red circles. The results of MF-PIFu of eight-view images looks better than four-view images. After voxel super-resolution, the coarse 3D models are refined to more accurate models, but the errors are not removed completely for the four-images. By contrast, the results of eight-view images look more smooth and accurate. Therefore, it is useful to obtain better estimation if there are more views. In this paper, it has been enough to obtain satisfying 3D models by four-view images. 
\begin{table}
\caption{Quantitative results for the four-view and eight-view images on the Articulated dataset.}
\label{tab:num_view}
\begin{tabular}{l|l|l|l}
\hline
View & P2S $\downarrow$  & Chamfer-$L_2$ $\downarrow$  & IoU $\uparrow$ \\
\hline
Ours(four views)& 0.3754 & 0.0032 & 0.9051 \\
Ours(Eight views)& \textbf{0.3606} & \textbf{0.0021} & \textbf{0.9042}\\
\hline
\end{tabular}
\end{table}
\section{Conclusion}
Detailed 3D human body reconstruction from RGB images is a challenging task because of the high number of degrees of the freedom of human body and the ambiguity of inferring 3D objects from 2D images. In this paper we propose a coarse-to-fine method for detailed 3D human body reconstruction from multi-view images through learning an implicit representation. The coarse 3D models are estimated from multi-view images through learning implicit representations based on multi-scale features which encode both local and global information. Then, generating the low-resolution voxel grids through voxelizing the coarse 3D models, we use voxel super-resolution to refine the coarse 3D models. For the voxel super-resolution, multi-stage 3D convolutional layers are used to extract multi-scale features from low-resolution voxel grids. The implicit representation is also learned based on the multi-scale features for voxel super-resolution. Benefiting from the voxel super-resolution, the coarse 3D models can be refined to have higher accuracy and better surface quality because the false reconstruction on the coarse 3D models can be removed and the details on the shape can be preserved. The experiments on the public datasets demonstrate that our method can recover detailed 3D human body models from multi-view images with higher accuracy and completeness than previous approaches. 

Some work needs to be done in the future. Firstly, we need to increase the variety of the training dataset. The models in the two datasets of our paper mostly have the same color clothes. If there is a new model with colorful clothes, our method will fail to obtain good results. However, the high-quality 3D human body models are not easy to be acquired and many datasets are not free, which increases the difficulty for the research. Besides, the texture of the detailed model is not considered in our method which should be done in the future. Finally, single-view image based reconstruction is needed in the future to increase the convenience of our method.
\bibliographystyle{spmpsci}      
\bibliography{egbib}   

%
%

\end{document}